\documentclass[10pt,journal,compsoc]{IEEEtran}
\ifCLASSOPTIONcompsoc
  \usepackage[nocompress]{cite}
\else
  \usepackage{cite}
\fi

\ifCLASSINFOpdf
  \usepackage[pdftex]{graphicx}
  \graphicspath{
    {./fig/}
  }
\else
\fi

\usepackage{amsmath}
\usepackage{amssymb}
\usepackage{color}
\usepackage{url}

\newcommand{\exten}{\textcolor{black}}
\newcommand{\lyj}{\textcolor{black}}
\newcommand{\YL}{\textcolor{black}}
\newcommand{\rev}{\textcolor{black}}
\newcommand{\lyjrev}{\textcolor{black}}
\newcommand{\minorrev}{\textcolor{black}}

\begin{document}
%
\title{\lyj{Quality Metric Guided Portrait Line Drawing Generation from Unpaired Training Data}}

\author{Ran Yi,
        Yong-Jin Liu,~\IEEEmembership{Senior Member,~IEEE,}
        Yu-Kun Lai,~\IEEEmembership{Member,~IEEE,}
        Paul L. Rosin
\IEEEcompsocitemizethanks{\IEEEcompsocthanksitem R. Yi is with BNRist, MOE-Key Laboratory of Pervasive Computing, the Department of Computer Science and Technology, Tsinghua University, Beijing, China; and the Department of Computer Science and Engineering, Shanghai Jiao Tong University, Shanghai, China.
\IEEEcompsocthanksitem Y.-J. Liu is with BNRist, MOE-Key Laboratory of Pervasive Computing, the Department of Computer Science and Technology, Tsinghua University, Beijing, China. Y.-J. Liu is the corresponding author. E-mail: liuyongjin@tsinghua.edu.cn.
\IEEEcompsocthanksitem Y.-K. Lai and P.L. Rosin are with School of Computer Science and Informatics, Cardiff University, UK.}
}

\markboth{Accepted by IEEE Transactions on Pattern Analysis and Machine Intelligence,
https://doi.org/10.1109/TPAMI.2022.3147570
}
{Yi \MakeLowercase{\textit{et al.}}: Bare Demo of IEEEtran.cls for Computer Society Journals}
%

\IEEEtitleabstractindextext{%
\begin{abstract}
\lyj{
Face portrait line drawing is a unique style of art which is highly abstract and expressive.
However, due to its high semantic constraints, many existing methods learn to generate portrait drawings using paired training data, which is costly and time-consuming to obtain.
In this paper, we propose a novel method to automatically transform face photos to portrait drawings using unpaired training data \lyjrev{with two new features; i.e.,} our method can (1) learn to generate high quality portrait drawings in multiple styles using a single network and (2) generate portrait drawings in a ``new style'' unseen in the training data. To achieve these benefits, we (1) propose a novel quality metric for portrait drawings which is learned from human perception, and (2) introduce a quality loss to guide the network toward generating better looking portrait drawings. \lyjrev{We observe that existing unpaired translation methods such as CycleGAN} tend to embed invisible reconstruction information indiscriminately in the whole drawings due to significant information imbalance between the photo and portrait drawing domains, which leads to important facial features missing. To address this problem, we propose a novel asymmetric cycle mapping that enforces the reconstruction information to be visible and only embedded in the \minorrev{selected} 
facial regions. 
Along with localized discriminators for important facial regions, our method well preserves all important facial features in the generated drawings.
Generator dissection further explains that our model learns to incorporate face semantic information during drawing generation.
Extensive experiments including a user study show that our model outperforms state-of-the-art methods.}
\end{abstract}

\begin{IEEEkeywords}
\exten{Face portrait, Drawing, Style transfer, Unpaired image translation, Generative adversarial network, Quality metric}
\end{IEEEkeywords}}

\maketitle

\IEEEdisplaynontitleabstractindextext

\IEEEpeerreviewmaketitle

\section{Introduction}
\label{sec:intro}

\rev{\IEEEPARstart{F}{ace} portrait line drawing is a highly abstract and expressive art form, which 
\minorrev{compresses}
the rich information in human portraits into a sparse set of graphical elements (e.g. lines) and has  high semantic constraints.}
Usually only skilled artists can generate delicate portrait line drawings and different artists have diverse styles. However, the hand-made drawing process is time consuming and 
challenging.
Recently, a few state-of-the-art works develop elegant algorithms to automatically generate face portrait line drawings \cite{YiLLR19,YiLLR20pami,YiLLR20}, which show some interesting progress on the aspect that artificial intelligence can learn to create human art. In this paper, we take a step forward by addressing the following problem: {\it Can artificial intelligence learn the artistic style space of face portrait line drawings and generate portrait drawings of ``new styles'' unseen in the training data?}

\begin{figure*}[thb]
\centering
\includegraphics[width = 1.0\textwidth]{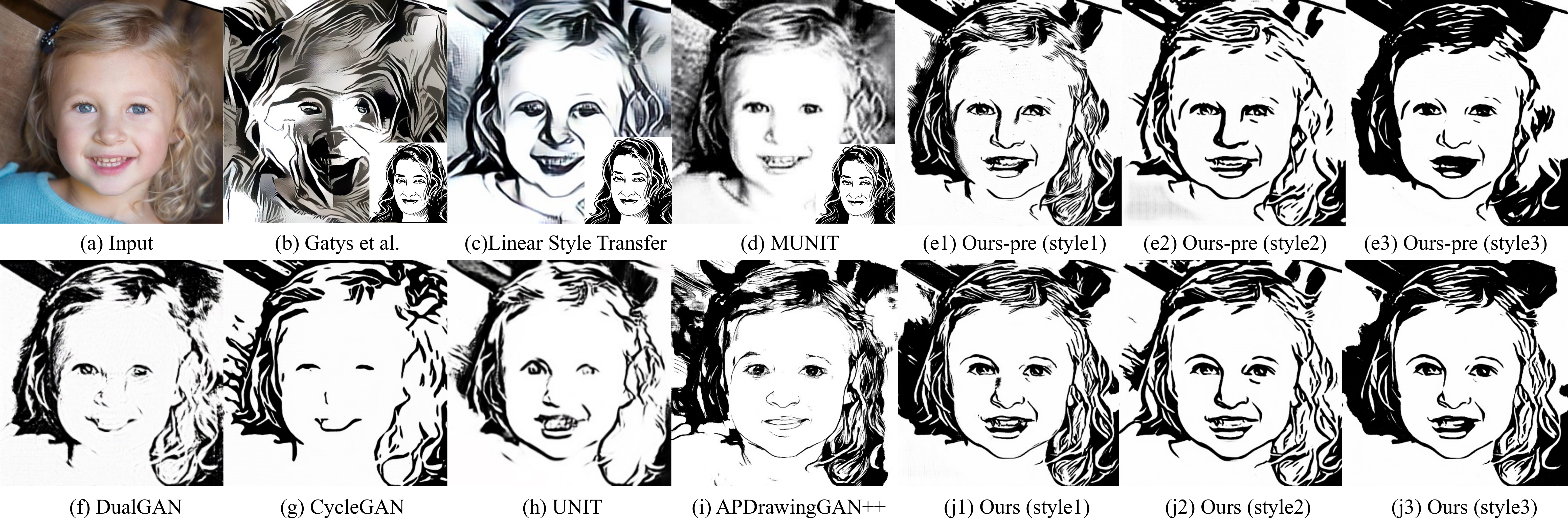}
\vspace{-0.2in}
\caption{\rev{Comparison with state-of-the-art methods: (a) input face photo; (b)-(c) style transfer methods: Gatys~\cite{GatysEB16} and Linear Style Transfer~\cite{LiLK019}; (f)-(h) single-modal image-to-image translation methods: DualGAN~\cite{YiZTG17}, CycleGAN~\cite{ZhuPIE17}, UNIT~\cite{LiuBK17}; (d) multi-modal image-to-image translation methods MUNIT~\cite{HuangLBK18}; (e) our previous conference version (Ours-pre) \cite{YiLLR20} that outputs three styles; (i) a portrait generation method APDrawingGAN++~\cite{YiLLR20pami} using paired training data; (j) our method. Note that our method only uses unpaired training data. Due to this essential difference, we only compare APDrawingGAN++ with our method in Appendix E.2.}}
\label{fig:teaser}
\vspace{-0.1in}
\end{figure*}

\lyj{This challenging problem has not been studied in previous research, possibly due to two outstanding issues. First, artistic  portrait  line  drawings  (APDrawings)  are  quite different from the previously tackled image styles.
Image style transfer has been a longstanding topic in computer vision.
In recent years, inspired by the effectiveness of deep learning, Gatys et al.~\cite{GatysEB16} introduced convolutional neural networks (CNNs) to transfer the style from a style image to a content image, and opened up the field of neural style transfer.
Subsequently, generative adversarial networks (GANs) have achieved much success in solving image style transfer problems~\cite{IsolaZZE17,ZhuPIE17}. However, existing methods are mainly applied to cluttered styles (e.g., oil painting style) where (1) a stylized image is full of fragmented brush strokes and (2) the requirement for the quality of each individual element is low. APDrawings are completely different\minorrev{,} and generating  them  is  very  challenging  because  the  style is highly  abstract:  it (1) only  contains  a  sparse  set  of  graphical  elements,  (2) is  line-stroke-based and disables shading, and (3) has high semantic constraints. Therefore, previous texture-based style transfer methods and  general image-to-image translation methods fail to generate good APDrawing results (Fig.~\ref{fig:teaser}).}

\lyj{The second outstanding issue is to use unpaired training data. The artistic
style space of APDrawings contains diverse styles and collecting paired training data for each style is impossible. APDrawingGAN \cite{YiLLR19} and APDrawingGAN++~\cite{YiLLR20pami} are the only methods that explicitly deal with APDrawings by using a hierarchical structure.
However, these methods require \emph{paired} training data that is costly to obtain.}
Compared to paired training data, APDrawing generation learned from \emph{unpaired} data is much more challenging.
Previous methods for unpaired image-to-image translation~\cite{ZhuPIE17,YiZTG17} use a cycle structure to regularize training.
Although cycle consistency loss enables learning from unpaired data, we observe that when applying them to face photo to APDrawing translation, due to significant imbalance of information richness in these two data types (\YL{accurately recovering a photo from the corresponding line drawing is an impossible task}), these methods tend to embed invisible reconstruction information indiscriminately in the whole APDrawing, causing a deterioration in the quality of the generated APDrawings, such as important facial features partially missing (Figs.~\ref{fig:teaser}(f-g)).

Our previous conference work~\cite{YiLLR20} partially addressed the unpaired training data issue by proposing an asymmetric cycle structure  \rev{and a truncation loss} to 
\rev{prevent the model from embedding invisible features in the generated APDrawings}.
\exten{In this paper, we substantially improve upon~\cite{YiLLR20}, and propose a \lyjrev{novel quality-metric-guided} \lyj{APDrawing generation model, which can generate (1) ``better looking'' APDrawings according to human perception, and (2) ``new style'' APDrawings other than the styles in the training data.
Learning from unpaired data makes our model able to utilize diverse APDrawing styles from web data for training the style space. To exploit the natural diversity of styles from web training images (see Fig.~\ref{fig:style_sample} for some examples), our model can (1) learn \emph{multiple styles} (as well as the style space) of APDrawings from \lyjrev{web data of mixed styles}, (2) generate ``new styles'' unseen in the training data, and (3) control the output style using a code in the style space.}} The source code is available\footnote{\url{https://github.com/yiranran/QMUPD}}.

\exten{
In particular, we make the following contributions:
\begin{itemize}
\item We propose a novel quality metric for \lyj{APDrawings} by learning from human perception. The new quality metric is modeled by a regression network whose input is \lyj{APDrawing} alone and the output is a quality score.
\item \lyjrev{Based on the quality metric, we propose a quality loss that is consistent with human perception, and use it to} guide the network toward generating better looking \lyj{APDrawings}.
\item We generate \lyj{APDrawings of} ``new styles'' unseen in the training data by searching for a corresponding style code in the style space.
\item \lyj{To interpret our model,} we dissect the generator by visualizing feature maps and comparing them to face semantics, \lyj{validating that our generator learns to incorporate semantic face information during APDrawing generation.}
\end{itemize}
}

\section{Related Work}

\subsection{Neural Style Transfer}

\lyj{Inspired by the successes of CNNs in many visual perception tasks, Gatys et al.~\cite{GatysEB16} proposed to use a pretrained CNN to transfer the style in an image to the content of another image in two steps. First, the content features and style features are extracted from images. Second, the content image is optimized by matching the style features from the style image \minorrev{while} simultaneously maintaining the content features. In \cite{GatysEB16}, the Gram matrix is used to measure style similarity.}
This method opens up the field of neural style transfer and many follow-up methods are proposed based on this.

\lyj{Li and Wand~\cite{LiW16} proposed to replace the Gram matrix by a Markov Random Field (MRF) regularizer for modeling the style. Stylized images are synthesized by combining MRF with CNN.
To speed up the slow optimization process of~\cite{GatysEB16}, some methods (e.g., \cite{JohnsonAF16,UlyanovLVL16})
use a feed-forward neural network to minimize the same objective function.
However, these methods still suffer from the problem that each model is restricted to a single style.
To simultaneously speed up optimization and maintain style flexibility as~\cite{GatysEB16}, Huang and Belongie~\cite{HuangB17} proposed adaptive instance normalization (AdaIN) to align the mean and variance of content features to those of style features.
In these example-guided style transfer methods, the style is extracted from a single image, which is not as convincing as learning from a set of images to synthesize style (refer to GAN-based methods in Section~\ref{ssec:relatedwork_gan}).}

\lyj{In principle, some neural style transfer methods can generate images with styles unseen in the training data (e.g., \cite{GatysEB16}). However, these methods model style as texture, and are not suitable for our APDrawing style that has little texture.}

\subsection{GAN-based image-to-image translation}
\label{ssec:relatedwork_gan}

GANs~\cite{GoodfellowPMXWOCB14} have achieved much progress in many \lyj{vision applications}, including image super-resolution~\cite{LedigTHCCAATTWS17}, text-to-image synthesis~\cite{ReedAYLSL16,ZhangXL17}, 
and facial attribute manipulation~\cite{ChoiCKH0C18},
etc.
Among these works, two unified GAN frameworks, Pix2Pix~\cite{IsolaZZE17} and CycleGAN~\cite{ZhuPIE17}, \lyj{have boosted the field of image-to-image translation.}

\begin{figure}[t]
\centering
\includegraphics[width = .49\textwidth]{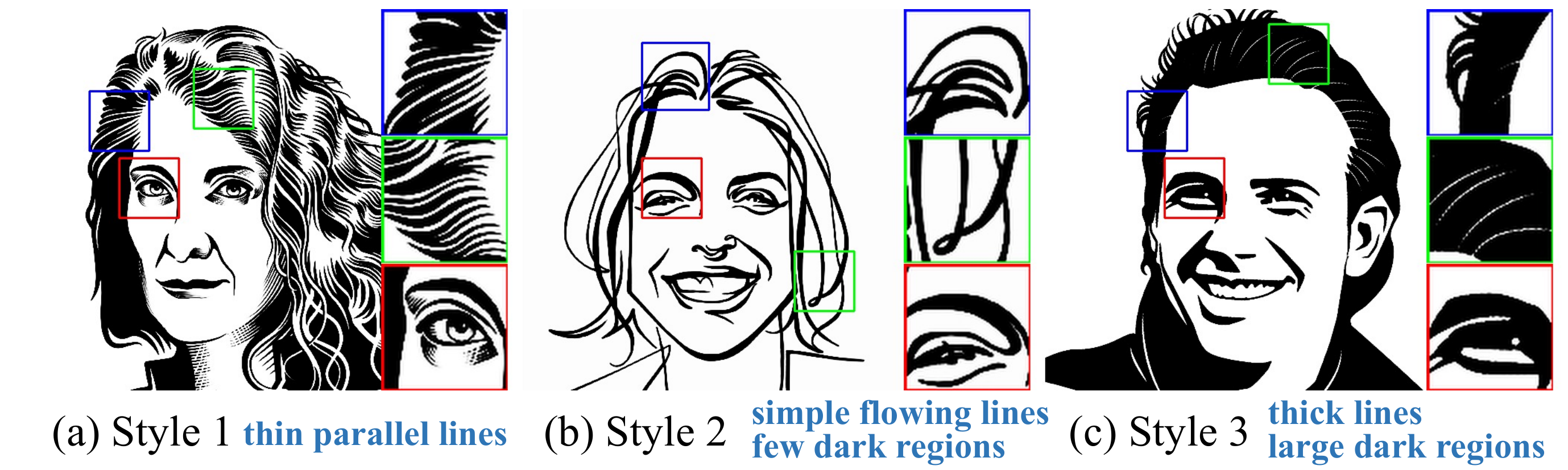}
\vspace{-0.2in}
\caption{\rev{\lyjrev{
\minorrev{Representative images of the three styles}
} in our collected web portrait line drawing data. (a) The first style is from Yann Legendre and Charles Burns who use \rev{thin} parallel lines to draw shadows. (b) The second style is from  Kathryn Rathke who draws facial features using simple flowing lines and uses few dark regions. (c) The third style is from {vectorportal.com} where continuous thick lines and large dark regions are utilized. Close up views are presented \minorrev{alongside} for better comparison of the three styles.}}
\label{fig:style_sample}
\vspace{-0.05in}
\end{figure}

Pix2Pix~\cite{IsolaZZE17} was the first general image-to-image translation framework based on conditional GANs, \lyj{and was later extended to high-resolution image synthesis~\cite{Wang0ZTKC18}. Pix2Pix is trained by paired data.
Recently, more works focus on learning from unpaired data, due to the difficulty of obtaining paired images in two domains. In this direction, two representative works are CycleGAN~\cite{ZhuPIE17} and DualGAN~\cite{YiZTG17}, which make use of the cycle consistency constraint.
This constraint enforces that the two mappings from domains $A$ to $B$ and from $B$ to $A$ --- when applied consecutively to an image --- revert the image back to itself.
Different from enforcing cycle consistency at the image level, UNIT~\cite{LiuBK17} tackles the problem by a shared latent space assumption and enforcing a feature-level cycle consistency.
These methods work well for general image-to-image translation tasks. However, in the transformation from face photos to APDrawings, we observe that cycle consistency constraints lead to facial features partially missing in APDrawings, because the information between the source and target domains is imbalanced. In this paper, we relax the cycle consistency in the forward cycle (i.e., photo $\rightarrow$ drawing $\rightarrow$ photo) and propose additional constraints to avoid this problem.
The NIR-to-RGB method in ~\cite{Dou0HP19} adopts a very different type of asymmetry --- it uses the same loss for the forward and backward cycles, and only changes the network complexity --- and targets a different task from ours.}

The aforementioned unpaired translation methods are also limited in the diversity of translation outputs.
Unpaired data such as crawled web data often naturally contains multi-modal distributions (i.e. inconsistent styles).
When knowing the exact number of modes and the mode each sample belongs to, the multi-modal image-to-image translation could be solved by treating each mode as a separate domain and using a multi-domain translation method~\cite{AnooshehATG18,ChoiCKH0C18}.
However, in many scenarios including our problem setting, this information is not available.
MUNIT~\cite{HuangLBK18} deals with multi-modal image-to-image translation without knowing the mode each sample belongs to.
It encodes an image into a domain-invariant content code and a domain-specific style code, and recombines the content code with the style code sampled from a target domain.
Although MUNIT generates multiple outputs with different styles, it cannot generate satisfactory portrait line drawings with clear lines.
\lyj{By inserting style features into the generator and using a soft classification loss to discriminate modes in the training data, our network architecture proposed in this paper  can produce multi-style outputs and generate better looking APDrawings than state-of-the-art methods.}

\section{\exten{Quality Metric for \lyj{APDrawings}}}
\label{sec:metric_model}

\begin{figure}[t]
\centering
\includegraphics[width = .5\textwidth]{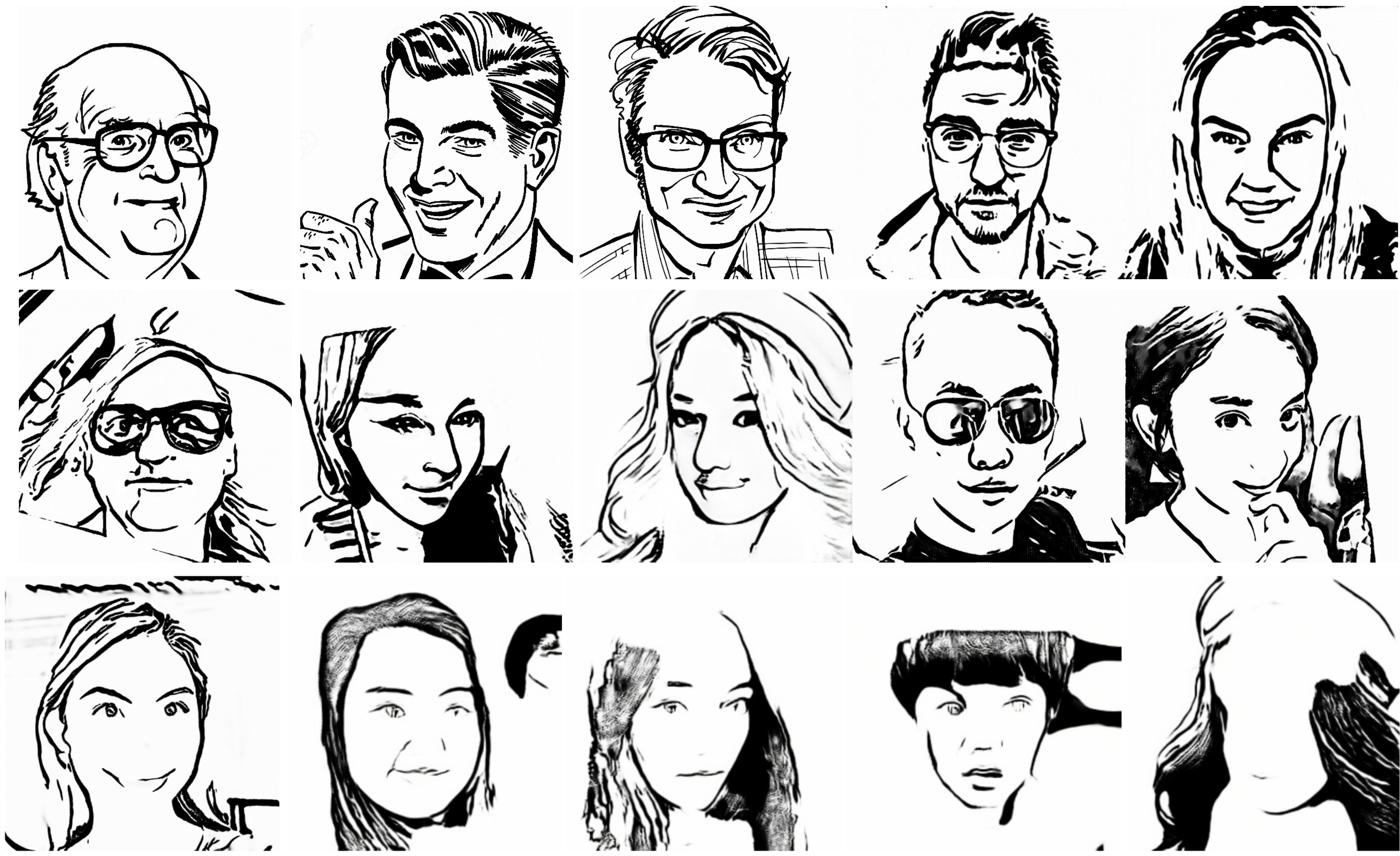}
\vspace{-0.15in}
\caption{\exten{Samples of collected (including generated and artist) portrait line drawings of target style 2 for quality metric model training. The drawings from top to bottom have decreasing quality.}}
\label{fig:data_prepare}
\vspace{-0.05in}
\end{figure}

\exten{Most image-to-image translation methods guide generation towards the target domain using a discriminator which decides whether an image is a real target-domain image or not.
When the target domain is \lyj{APDrawings}, we found it is not sufficient to decide whether such a drawing is real or fake; \lyj{i.e., the generator needs further to be told the quality of the synthesized drawing. 
To the best of our knowledge, there lacks an optimization tool} to encourage the network to generate \emph{good looking} portrait line drawings. 
\lyjrev{The perception of good APDrawings} --- e.g., fluent lines and avoiding messy lines on the face --- can be easily concluded by a human, but has not been \lyj{fully} described in an optimization goal.
Thus, we introduce a new quality metric for portrait line drawings by \emph{learning from human preference}.}

\exten{From previous user studies, we found that people can easily decide the quality of a portrait line drawing without knowing the original face photo. So our desired metric can be modeled by a regression network whose input is \lyj{an APDrawing alone} and the output is its quality score.}

\exten{To obtain such a regression network to predict APDrawing quality, we first generate many APDrawings using existing methods and mix these generated drawings with real artist drawings.
Then a user study is conducted to collect human preferences of these APDrawings. After calculating the quality score of each drawing based on human preference, we train a regression network to predict the quality score of an APDrawing.}

\subsection{Data Preparation}
\exten{We use existing unpaired image-to-image generation methods including DualGAN~\cite{YiZTG17}, CycleGAN~\cite{ZhuPIE17}, UNIT~\cite{LiuBK17}, ComboGAN~\cite{AnooshehATG18}, DRIT~\cite{LeeTHSY18} and our previous conference work~\cite{YiLLR20} to learn the three target styles (as specified in Fig.~\ref{fig:style_sample}).
We then test the trained models on collected web face photos and generate portrait line drawings for the three target styles.
\lyj{The generated drawings} are mixed with high-quality APDrawings for subsequent human evaluation.
\lyj{To facilitate the development of a good quality prediction model,} we include drawings with diverse quality in the data (as shown in Fig.~\ref{fig:data_prepare}).}

\subsection{Human Preference Collection and \lyj{Ranking}}

\exten{Considering pairwise comparison is one of the most practical and reliable methods to compare different \lyj{results}, we design a user study based on pairwise comparison. \lyj{From the comparison results, we compute a ranking.}
However, \lyj{given $n$ results,} \YL{obtaining} all $n(n-1)/2$ comparisons from a user study is time consuming and not feasible when $n$ is large.
We seek to utilize as few pairwise comparisons as possible to get a global ranking.
The efficient ranking method \lyj{in~\cite{WauthierJJ13}} is suitable in our application. It randomly conducts pairwise comparisons and estimates the score of an object as the relative difference of numbers of preceding items and succeeding items.
By using this efficient ranking strategy, an average of $O(n \log n)$ pairwise comparisons are sufficient to recover the true ranking.}

\exten{
{\bf User study design.}
Noting that it is difficult to compare the quality of portrait line drawings of different styles due to the style distractions, we \minorrev{therefore} only enable pairwise comparison between portrait line drawings of the same style.
Then we adopt the above efficient ranking strategy and conduct three user studies based on pairwise comparison for three target styles.
To simplify the answering process, each user is shown three portrait line drawings of the same style in a question and asked to rank the three drawings (\lyj{the answer to each question} equals to three pairwise comparisons). To balance the data amount and the effort for human evaluation, we randomly choose 250 drawings for each of the three target styles and collect $2,450 \sim 3,450$ question \minorrev{responses} for each style. 
}

\exten{
{\bf Score and ranking calculation.} 
We calculate the scores and global rankings for portrait line drawings of each style separately.
To compute the relative difference of numbers of preceding items and succeeding items \lyj{in}~\cite{WauthierJJ13}, for each question answer, denote the ranking as $I_1 \prec I_2 \prec I_3$, then the score of $I_1$ decreases by 2 (0 preceding and 2 succeeding), the score of $I_3$ increases by 2 (2 preceding and 0 succeeding) and the score of $I_2$ stays unchanged.
By summarizing all question \minorrev{responses for} a style, we calculate the score for each drawing of this style and get a global ranking based on the score.
The scores are then normalized to the range \lyj{$[0.1,1]$\footnote{\minorrev{The lower bound is greater than zero since even the worst examples of training data are better than random.}} so that drawing scores of the three styles have the same range.}
}

\subsection{Quality Metric Prediction}
\exten{Given the portrait drawing data and the normalized quality score, we train a regression network to predict \lyj{APDrawing} quality.
The regression network is based on the Inception~v3~\cite{SzegedyVISW16} architecture.
It takes an APDrawing as input and outputs a quality value.
We gather the drawing data of three target styles and train a unified prediction model $M$.
}

\exten{Since the quality metric model behavior is learned from human evaluation, the predicted score can help guide the drawing generator toward better quality. 
\lyjrev{Furthermore, it can also be used to choose which trained version of the photo-to-APDrawing generator to use, e.g. when multiple versions are trained using different hyper-parameters}.
}

\begin{figure}[tb]
\centering
\includegraphics[width = 0.9\columnwidth]{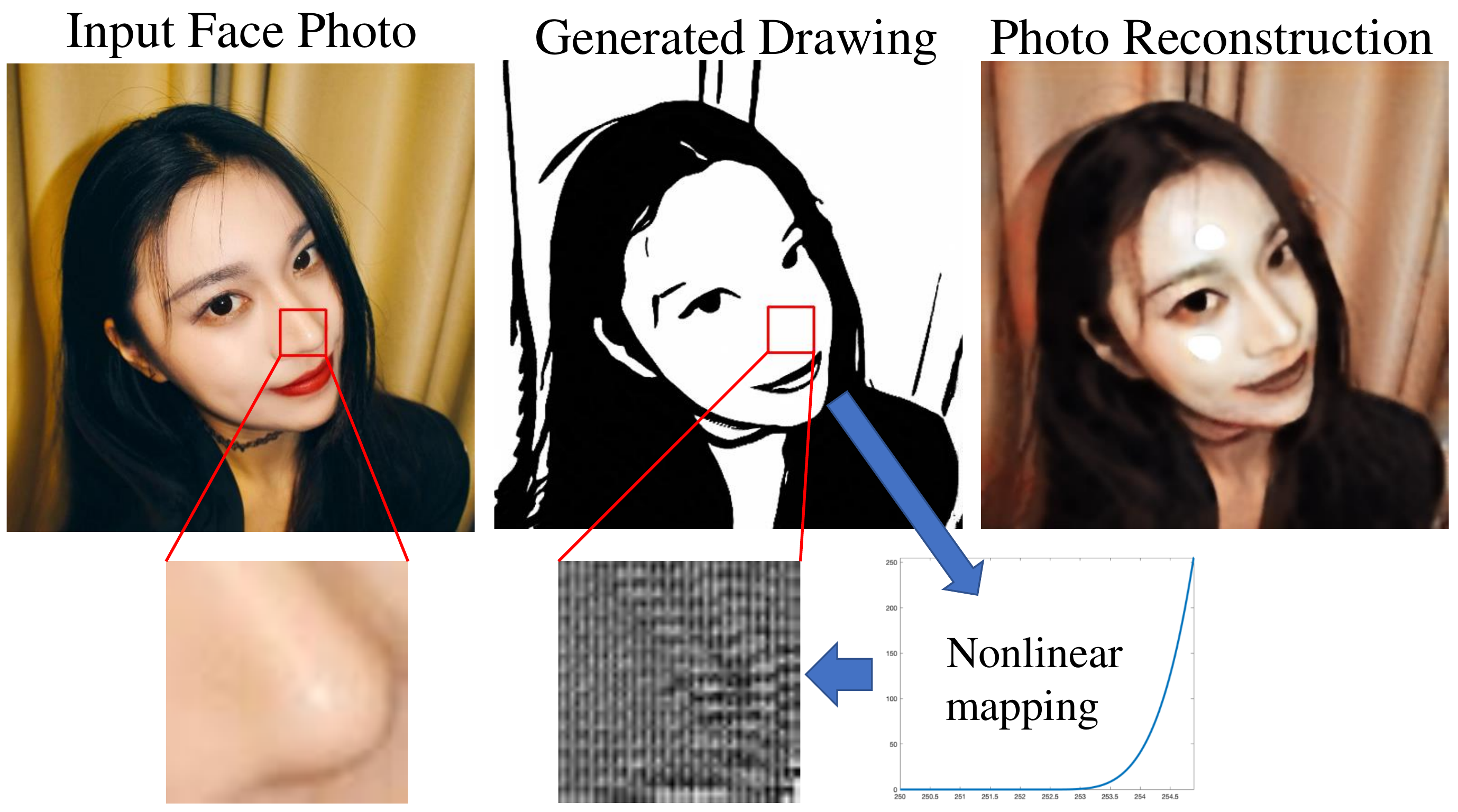}
\caption{\lyjrev{CycleGAN reconstructs the input photo from generated drawings using a strict cycle-consistency loss, which 
\minorrev{can potentially embed invisible reconstruction information anywhere}
in the whole drawings.} A nonlinear monotonic mapping of the gray values is applied in a local region around the nose to visualize the embedded reconstruction information.}
\label{fig:hideinfo}
\end{figure}

\section{\lyj{Network Architecture and \lyj{Optimization}}}

\subsection{Overview}

\lyj{With the aid of trained quality metric model (Section~\ref{sec:metric_model}), in this section, we propose a GAN model with a novel asymmetric cycle structure, that transforms face photos to high-quality APDrawings, only using unpaired training data.}
Let $\mathcal{P}$ and $\mathcal{D}$ be the face photo domain and the APDrawing domain, and no pairings need to exist between these two domains.
Our model learns a function $\Phi$ that maps from $\mathcal{P}$ to $\mathcal{D}$ using training data \lyj{$S(p) = \{p_i | i = 1,2,\cdots,n_p\} \subset \mathcal{P}$ and $S(d) = \{d_j | j = 1,2,\cdots,n_d\} \subset \mathcal{D}$. $n_p$ and $n_d$ are the numbers of training photos and APDrawings.}
\lyjrev{Our model consists of (1) two generators, i.e., a generator $G$ transforming face photos to APDrawings, and an inverse generator $F$ transforming APDrawings back to face photos, and (2) two discriminators, i.e., $D_{\mathcal{D}}$ responsible for discriminating generated drawings from real drawings, and $D_{\mathcal{P}}$ responsible for discriminating generated photos from real photos.}

The information in the APDrawing domain is much less than in the face photo domain. For example, in the cheek region, there are many color variations in the original photo but the cheek is usually drawn completely white (i.e. no lines are included) in an APDrawing. \lyj{As illustrated in Fig. \ref{fig:hideinfo},}
enforcing a strict cycle-consistency loss like in CycleGAN~\cite{ZhuPIE17} on \lyj{the reconstructed and input photos} will cause the network to embed reconstruction information in very small variations in the generated APDrawings (i.e., color changes that are invisible to the eye but can make a difference in network calculation); \lyj{a similar phenomenon was observed in~\cite{abs-1712-02950}.}
Embedding reconstruction information in very small variations achieves a balance between cycle-consistency loss and GAN loss in CycleGAN; \lyj{from the generated drawing $G(p)$, a face photo similar to the input photo $p$ can be successfully reconstructed  because of small color changes,} while at the same time $G(p)$ \lyj{tries to} be similar to real drawings and be classified as real by the discriminator. \lyj{However, in APDrawing generation,}
embedding invisible reconstruction information indiscriminately in the whole drawing will put a strong restriction on the objective function optimization. Moreover, it will allow important facial features to be partially missing in the generated drawings.

\begin{figure*}[thb]
\centering
\includegraphics[width = 0.88\textwidth]{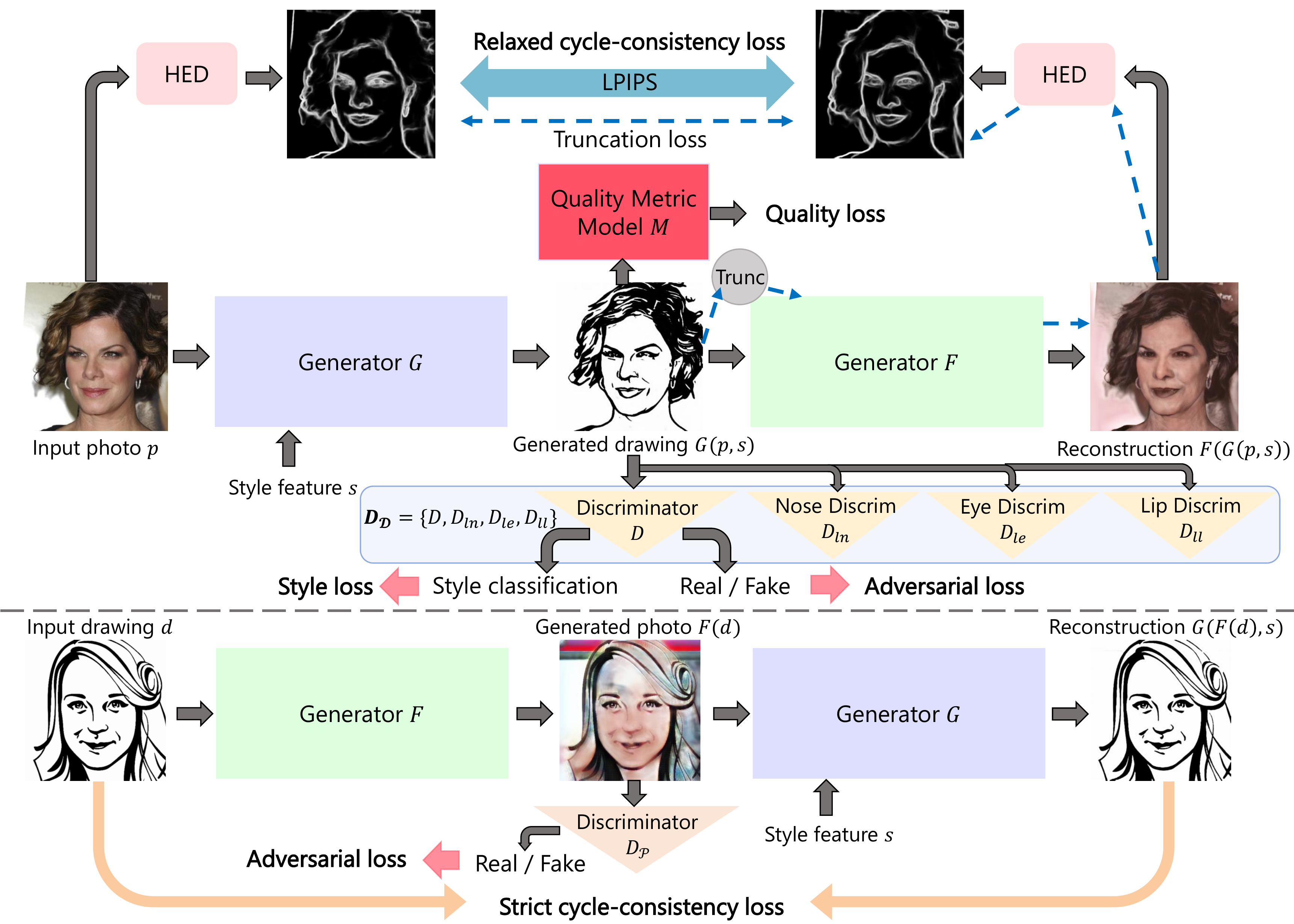}
\vspace{-0.05in}
\caption{\lyj{Our GAN model uses an asymmetric cycle structure, which} consists of a photo to drawing generator $G$, a drawing to photo generator $F$, a drawing discriminator $D_{\mathcal{D}}$ and a photo discriminator $D_{\mathcal{P}}$. We use a relaxed cycle-consistency loss between reconstructed face photo \rev{$F(G(p,s))$} and input photo $p$, while enforcing a strict cycle-consistency loss between reconstructed drawing $G(F(d))$ and input drawing $d$. We \lyjrev{also} introduce local drawing discriminators $D_{ln},D_{le}, D_{ll}$ for the nose, eyes and lips
and a truncation loss. Our model deals with multi-style generation by inserting a style feature \lyj{vector} into the generator and adding a style loss. \exten{A quality loss \lyj{based on the quality metric model (Section \ref{sec:metric_model})} further encourages generation of ``good looking'' APDrawings.} \lyj{The detailed architecture is illustrated in Appendix B.}}
\label{fig:framework}
\end{figure*}

\lyj{To address this problem, our model utilizes a novel idea:} although cycle consistency constraints are useful to regularize training, we are only interested in the one way mapping from photos to \lyj{APDrawings}.
Therefore, unlike CycleGAN, we do not expect or require the inverse generator $F$ to reconstruct a face photo exactly as the input photo (which is a near impossible task).
Instead, our proposed model is \emph{asymmetric} in that we use a relaxed cycle-consistency loss between $F(G(p))$ and $p$, where only edge information is enforced to be similar, while a strict cycle-consistency loss is enforced on $G(F(d))$ and $d$.
By tolerating the reconstruction information loss between $F(G(p))$ and $p$, the objective function optimization has enough flexibility to recover all important facial features in APDrawings.
A truncation loss is further proposed to enforce the embedded information to be visible, especially around the local area of the selected edges where relaxed cycle-consistency loss works.
Furthermore, local drawing discriminators for the nose, eyes and lips are introduced to enforce \lyj{the significance of} their existence and ensure quality for these regions in the generated drawings. \lyj{By integrating these techniques, our model can generate high-quality APDrawings with complete facial features.}

\lyj{Another benefit of our model is to generate multi-style APDrawings.}
The APDrawing data we collected from the Internet contains a variety of styles, of which only some are tagged with author/source information.
We select representative styles from the collected data (Fig.~\ref{fig:style_sample}), and train a classifier for the collected drawings.
Then a learned representation is extracted as a style feature \lyj{vector} and inserted into the generator to control the generated drawing style.
An additional style loss is introduced to optimize for each style.

\lyj{Different from our previous conference work \cite{YiLLR20},} \exten{we further improve the APDrawing quality and alleviate unwanted artifacts by utilizing the trained quality metric model (Section~\ref{sec:metric_model}).
A \lyj{human-perception-consistent} quality metric loss is proposed to guide the network toward generating \lyj{good looking APDrawings.}
}

The four networks in our model are trained in an adversarial manner~\cite{GoodfellowPMXWOCB14}: \lyj{(1)} the two discriminators $D_{\mathcal{D}}$ and $D_{\mathcal{P}}$ are trained to maximize the probability of assigning correct labels to real and synthesized drawings and photos,
and \lyj{(2)} meanwhile the two generators $G$ and $F$ are trained to minimize the probability of the discriminators assigning the correct labels.
The loss function $L(G,F,D_{\mathcal{D}},D_{\mathcal{P}})$ contains six types of loss terms: adversarial loss $L_{adv}(G,D_{\mathcal{D}})+L_{adv}(F,D_{\mathcal{P}})$, relaxed cycle consistency loss $L_{relaxed\sim cyc}(G,F)$, strict cycle consistency loss $L_{strict\sim cyc}(G,F)$, truncation loss $L_{trunc}(G,F)$, style loss $L_{style}(G,D_{\mathcal{D}})$\exten{, and quality loss based on the quality metric model $L_{quality}(G)$}.
Then the function $\Phi$ is optimized by solving the minimax problem with \lyj{the} loss function: 
\begin{equation}
\begin{aligned}
&\min_{G,F}\max_{D_{\mathcal{D}},D_{\mathcal{P}}}L(G,F,D_{\mathcal{D}},D_{\mathcal{P}}) \\
&=(L_{adv}(G,D_{\mathcal{D}})+L_{adv}(F,D_{\mathcal{P}})) \\
&+ \lambda_1 L_{relaxed\sim cyc}(G,F) + \lambda_2 L_{strict\sim cyc}(G,F) \\
&+ \lambda_3 L_{trunc}(G,F) + \lambda_4 L_{style}(G,D_{\mathcal{D}}) + \lambda_5 L_{quality}(G)
\end{aligned}
\label{eq:loss}
\end{equation}

\lyj{The network architectures for $G$, $F$, $D_{\mathcal{D}}$ and $D_{\mathcal{P}}$ are introduced in Section~\ref{ssec:arch}.
The detailed design of six loss terms are presented in Section~\ref{ssec:loss}. An overview of our model is illustrated in Fig.~\ref{fig:framework}.}

\subsection{\lyj{Network} Architecture}
\label{ssec:arch}

\lyj{Our GAN model consists of (1) a generator $G$ and a drawing discriminator $D_{\mathcal{D}}$ for face photo to APDrawing translation, and (2) another generator $F$ and a photo discriminator $D_{\mathcal{P}}$ for the inverse APDrawing to photo translation.}
Considering information imbalance between the face photo in $\mathcal{P}$  and the APDrawing in $\mathcal{D}$, we design different architectures for $(G,D_{\mathcal{D}})$ and $(F,D_{\mathcal{P}})$.

\subsubsection{Face photo to \lyj{APDrawing} generator $G$}
\label{sssec:archg}
The generator $G$ takes a face photo $p$ and a style feature $s$ as input, and outputs an APDrawing $G(p,s)$ whose style is specified by $s$.

{\bf Style feature \lyj{vector} $s$.} We first train a classifier $C$ (based on VGG-19~\cite{SimonyanZ14a}) that classifies \lyj{APDrawings} into three styles (Fig.~\ref{fig:style_sample}), using tagged web drawing data.
Then we extract the output of the last fully-connected layer and use a softmax layer to calculate a 3-dimensional vector as the style feature for each drawing (including untagged ones).

{\bf Network structure.} $G$ is an encoder-decoder with residual blocks~\cite{HeZRS16} in the middle.
It starts with a flat convolution and two down convolution blocks to encode face photos and extract useful features.
Then the style feature is mapped to a 3-channel feature map and inserted into the network by concatenating it with the feature map of the second down convolution block.
An additional flat convolution is used to merge the style feature map with the extracted feature map.
Afterwards, nine residual blocks of the same structure are used to construct the content feature and transfer it to the target domain.
Then the output drawing is reconstructed by two up convolution blocks and a final convolution layer.

\subsubsection{Drawing discriminator $D_{\mathcal{D}}$}

The drawing discriminator $D_{\mathcal{D}}$ has two tasks: \lyj{1) discriminating generated APDrawings from real ones; and 2) classifying an APDrawing into three selected styles,} where a real drawing $d$ is expected to be classified into the correct style label (given by $C$), and a generated drawing $G(p,s)$ is expected to be classified into the style specified by the 3-dimensional style feature $s$.

For the first task, to enforce the existence of important facial features in the generated drawing, \lyj{in addition to a global discriminator $D$ that examines the full drawing,} we add three local discriminators $D_{ln},D_{le},D_{ll}$
to focus on discriminating \lyj{local drawings around the nose, eyes and lip respectively.} The inputs to these local discriminators are masked drawings, where masks are obtained from a face parsing network~\cite{GuBY0WY19}. \lyj{Finally}
$D_{\mathcal{D}}$ consists of $D,D_{ln},D_{le}$ and $D_{ll}$.

{\bf Network structure.} The global discriminator $D$ is based on PatchGAN~\cite{IsolaZZE17} and modified to have two branches.
The two branches share three
down convolution blocks. Then one branch $D_{rf}$ includes two flat convolution blocks to output a prediction map of real/fake for each patch in the drawing. \lyj{The} other classification branch $D_{cls}$ includes more down convolution blocks and outputs probability values for three style labels.
Local discriminators $D_{ln},D_{le},D_{ll}$ also adopt the PatchGAN structure.

\subsubsection{\lyj{APDrawing} to face photo generator $F$ and Photo discriminator $D_{\mathcal{P}}$}
The generator $F$ in the inverse direction takes \lyj{an APDrawing} $d$ as input and outputs a face photo $F(d)$. It adopts an encoder-decoder architecture with nine residual blocks in the middle.
Photo discriminator $D_{\mathcal{P}}$ discriminates generated face photos from real ones, and also adopts the PatchGAN structure.

\subsection{Loss Function}
\label{ssec:loss}
There are six types of losses
in our loss function (Eq. (\ref{eq:loss})). We explain them in detail as follows.

{\bf Adversarial loss.} The adversarial loss judges discriminator $D_{\mathcal{D}}$'s ability to assign correct labels to real and synthesized drawings.
It is formulated as:
\begin{equation}
\begin{aligned}
L_{adv}(G,D_{\mathcal{D}}) &= \sum_{D\in D_{\mathcal{D}}} \mathbb{E}_{d \in S(d)} [\log D(d)] \\
&+ \sum_{D\in D_{\mathcal{D}}} \mathbb{E}_{p \in S(p)} [\log (1-D(G(p,s))]
\end{aligned}
\end{equation}
where $s$ is randomly selected from the style feature \lyj{vectors of APDrawings in the training data $S(d)$} for each $p$.
As $D_{\mathcal{D}}$ maximizes this loss and $G$ minimizes it, this loss drives the generated drawings to become closer to real drawings.

We also adopt an adversarial loss for the photo discriminator $D_{\mathcal{P}}$ and the inverse mapping $F$:
\begin{equation}
\begin{aligned}
L_{adv}(F,D_{\mathcal{P}}) &= \mathbb{E}_{p \in S(p)} [\log D_{\mathcal{P}}(p)] \\
&+ \mathbb{E}_{d \in S(d)} [\log (1-D_{\mathcal{P}}(F(d))]
\end{aligned}
\end{equation}

{\bf Relaxed forward cycle-consistency loss.}
As \minorrev{previously} mentioned, we observe that there is much less information in domain $\mathcal{D}$ than information in domain $\mathcal{P}$.
\lyj{We do not expect the result from $p \rightarrow G(p,s) \rightarrow F(G(p,s))$ to be pixel-wise similar to $p$.}
Instead, we only expect the edge information in $p$ and $F(G(p,s))$ to be similar.
We extract edges from $p$ and $F(G(p,s))$ using HED~\cite{XieT15}, and evaluate the similarity of edges by the LPIPS perceptual metric proposed in~\cite{ZhangIESW18}.
Denote HED by $H$ and the perceptual metric by $L_{lpips}$. The relaxed cycle-consistency loss is formulated as:
\begin{equation}
\begin{aligned}
L_{relaxed\sim cyc}(G,F) &= \mathbb{E}_{p \in S(p)} [L_{lpips}(H(p),H(F(G(p,s))))]
\end{aligned}
\end{equation}

\begin{figure*}[t]
\centering
\includegraphics[width = 1.0\textwidth]{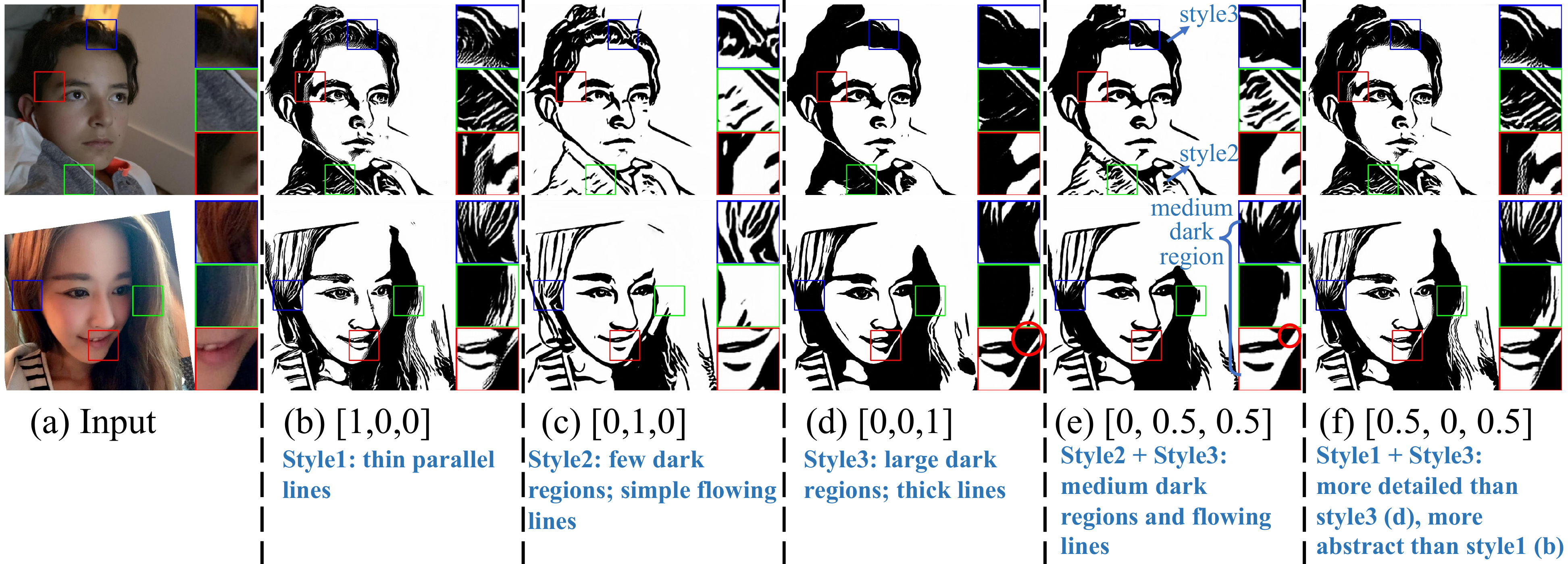}
\vspace{-0.2in}
\caption{\rev{Results of interpolating between style feature vectors: (a) input photos, (b)-(d) results of three target styles, (e)-(f) results of interpolating target styles. Close-up views are shown by the side.}}
\vspace{-0.15in}
\label{fig:style_interpolate}
\end{figure*}

{\bf Strict backward cycle-consistency loss.}
On the other hand, the information in the generated face photo is adequate to reconstruct the APDrawing.
Therefore, \lyj{we expect the result from $d \rightarrow F(d) \rightarrow G(F(d),s(d))$ to be pixel-wise similar to $d$, here $s(d)$ is the style feature of $d$.}
The strict cycle-consistency loss in the backward cycle is then formulated as:
\begin{equation}
\begin{aligned}
L_{strict-cyc}(G,F) &= \mathbb{E}_{d \in S(d)} [||d - G(F(d),s(d))||_1]
\end{aligned}
\end{equation}

{\bf Truncation loss.}
The truncation loss is designed to prevent the generated drawing from hiding information in small values.
It is in the same format as the relaxed cycle-consistency loss, \lyj{except} that the generated drawing $G(p,s)$ is first truncated to 6 bits\footnote{\lyj{Generally the intensity of a digital image is stored in 8 bits.}} to ensure \lyj{that} encoded information is clearly visible,
and then fed into $F$ to reconstruct the photo.
\rev{More specifically, we first scale the intensities to the range $[0, 64)$, truncate the
fractional part,
and then scale back.}
Denote the truncation operation as $T[\cdot]$, the truncation loss is formulated as:
\begin{equation}
\begin{aligned}
L_{trunc}(G,F) &= \mathbb{E}_{p \in S(p)} [L_{lpips}(H(p),H(F(T[G(p,s)])))]
\end{aligned}
\end{equation}
\lyj{At the beginning of training process,} the weight for the truncation loss is kept low, \lyj{otherwise it would be hard to optimize the model.}
The weight gradually increases as the training progresses.

{\bf Style loss.}
\lyj{It} is introduced to help $G$ generate multiple styles with different style features.
Denote the classification branch in $D_{\mathcal{D}}$ as $D_{cls}$. The style loss is formulated as
\begin{equation}
\begin{aligned}
\rev{L_{style}(G,D_{\mathcal{D}})} &= \mathbb{E}_{d \in S(d)} [-\sum_{c} p(c) \log D_{cls} (c | d)] \\
&+ \mathbb{E}_{p \in S(p)} [- \sum_{c} p'(c) \log D_{cls} (c | G(p,s))]
\end{aligned}
\end{equation}
For \lyj{a} real drawing $d$, $p(c)$ is the probability over style label $c$ given by classifier $C$, $D_{cls} (c | d)$ is the predicted softmax probability by $D_{cls}$ over $c$.
We multiply \YL{the term} by the probability $p(c)$ in order to take into account those real drawings that may not belong to a single style but lie between two styles, e.g. \YL{with softmax probabilities} $[0.58,0.40,0.02]$.
For generated drawing $G(p,s)$, $p'(c)$ denotes the probability over style label $c$ and is specified by style feature $s$, \lyj{and} $D_{cls} (c | G(p,s))$ is the predicted softmax probability over $c$.
This classification loss drives $D_{cls}$ to classify a drawing into the correct style and drives $G$ to generate a drawing close to a given style feature.

\exten{{\bf Quality loss based on the quality metric model.}
\lyj{It is designed for generating ``good looking'' APDrawings.
The quality metric model $M$ (described in Section~\ref{sec:metric_model}) predicts a quality score of an APDrawing by how consistent it is} with human perception, where better looking drawings get higher prediction scores (in the range \lyj{$[0.1,1]$}). We then define the quality loss $L_{quality}$ as
\begin{equation}
L_{quality}(G) = \mathbb{E}_{p \in S(p)}[1-M(G(p,s))].
\end{equation}
}

\section{\exten{New Style Generation}}

\lyj{In this section, we propose a solution to address the \lyjrev{challenging} problem of how to generate high-quality APDrawings of ``new styles'' unseen in the training data.}
\exten{In our multi-style generation setting, different style \lyj{feature} vectors lead to different style outputs.
The three target styles correspond to \lyj{vectors} $[1,0,0], [0,1,0], [0,0,1]$, respectively.
\lyj{An interesting} question is {\it \lyjrev{what results} other style \lyj{feature} vectors would generate and whether some style \lyj{feature} vectors could generate new styles unseen in the training data.}
}

\rev{\lyjrev{More specifically, the three target styles we used here  are representative styles of portrait line drawings, as shown in Fig.~\ref{fig:style_sample}.} These three styles vary in line thickness, arrangement and dark region ratio, etc. The \lyjrev{key features in the three styles are: 1) the drawings of style1 often use thin parallel lines to draw shadows, 2) the drawings of style2 use simple flowing lines and few dark regions, and 3) the drawings of style3 use thick lines and large dark regions.}}

\begin{figure*}[t]
\centering
\includegraphics[width = 1.0\textwidth]{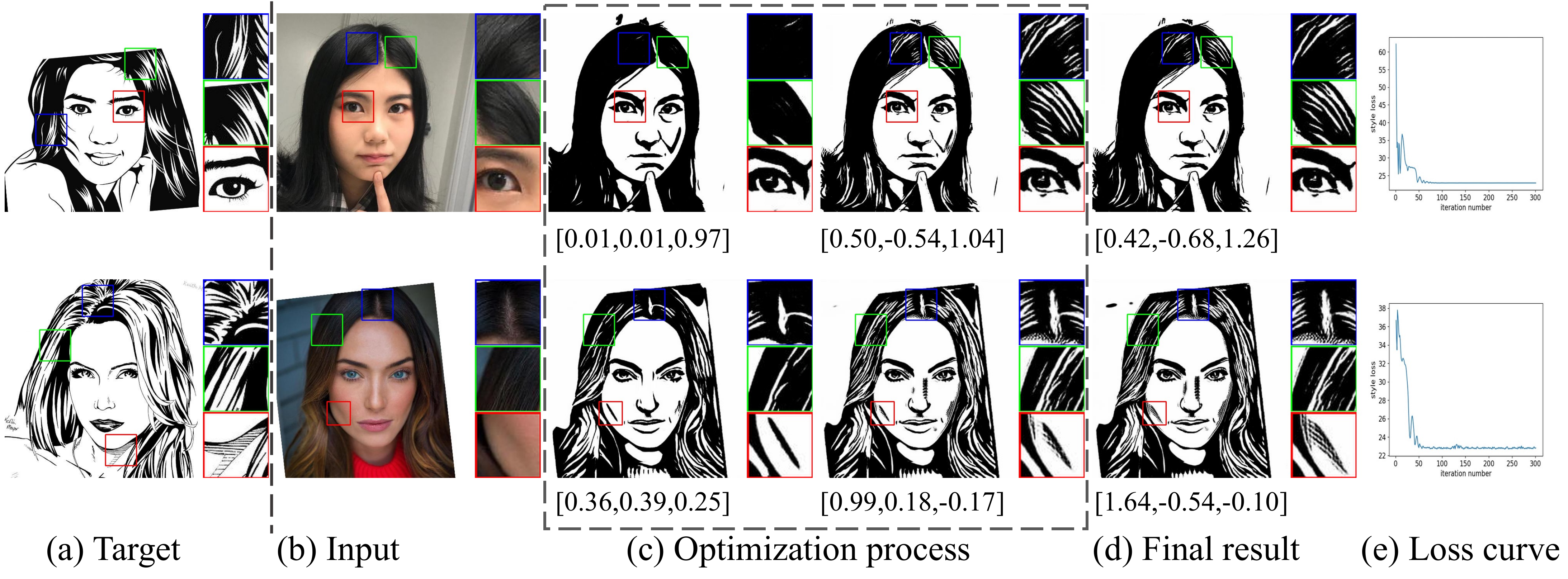}
\vspace{-0.3in}
\caption{\rev{Examples of ``new'' style generation. Given a target ``new'' style portrait drawing (i.e., style unseen in training data) in (a), we find a proper style feature vector that generates APDrawings similar to the target, by optimizing a histogram based style loss. The optimization process is shown in (c) and the final generated APDrawing is shown in (d). The style loss change during optimization is shown in (e). Style feature vectors used for generation are shown under each generated APDrawing. Close-up views are shown by the side.}}
\label{fig:style_optim}
\vspace{-0.05in}
\end{figure*}

\exten{By interpolating between the style feature vectors, we observe that the generated results show a combination of target styles. 
As shown in \rev{Fig.~\ref{fig:style_interpolate}e}, the results of style feature vector $[0, 0.5, 0.5]$ exhibit a combination of styles 2 and 3, i.e., medium dark regions and flowing lines; in other words, less dark regions compared to (d) and more compared to (c). 
The result of vector $[0.5, 0, 0.5]$ in \rev{Fig.~\ref{fig:style_interpolate}f} shows a combination of styles 1 and 3, i.e., a combination of thin parallel lines and dark region shadows; in other words, hair regions are more detailed (parallel lines) than (d) and more abstract than (b).
\rev{Close-up views are also provided for better comparison.}}

\exten{Next we explore whether the network can generate some ``new'' styles\footnote{\exten{Here, a ``new'' style portrait drawing means that the style is not one of the three target styles and unseen in the training data.}} unseen in the training data.
\lyj{Given a ``new'' style APDrawing $d_{target}$ as a reference, we use the trained APDrawing generator $G$ to look for a style feature vector $s$ in the style space that generates APDrawings most similar in style to the unseen target $d_{target}$.
The best style feature vector $s^*$ is found by optimizing the style distance between the generated APDrawing guided by this vector and the target $d_{target}$.}
Denote the loss term to measure style distance as $L_{style}$. The problem is formulated as:
\begin{equation}
s^*=\arg \min_s L_{style}(G(p,s), d_{target})
\label{eq:vec_op}
\end{equation}
where $p$ is a face photo in the training data.
Examples of ``new'' style generation and the corresponding ``new'' style targets are presented in Figures~\ref{fig:style_optim}(a-d).
}

\begin{figure*}[t]
\centering
\includegraphics[width = 0.9\textwidth]{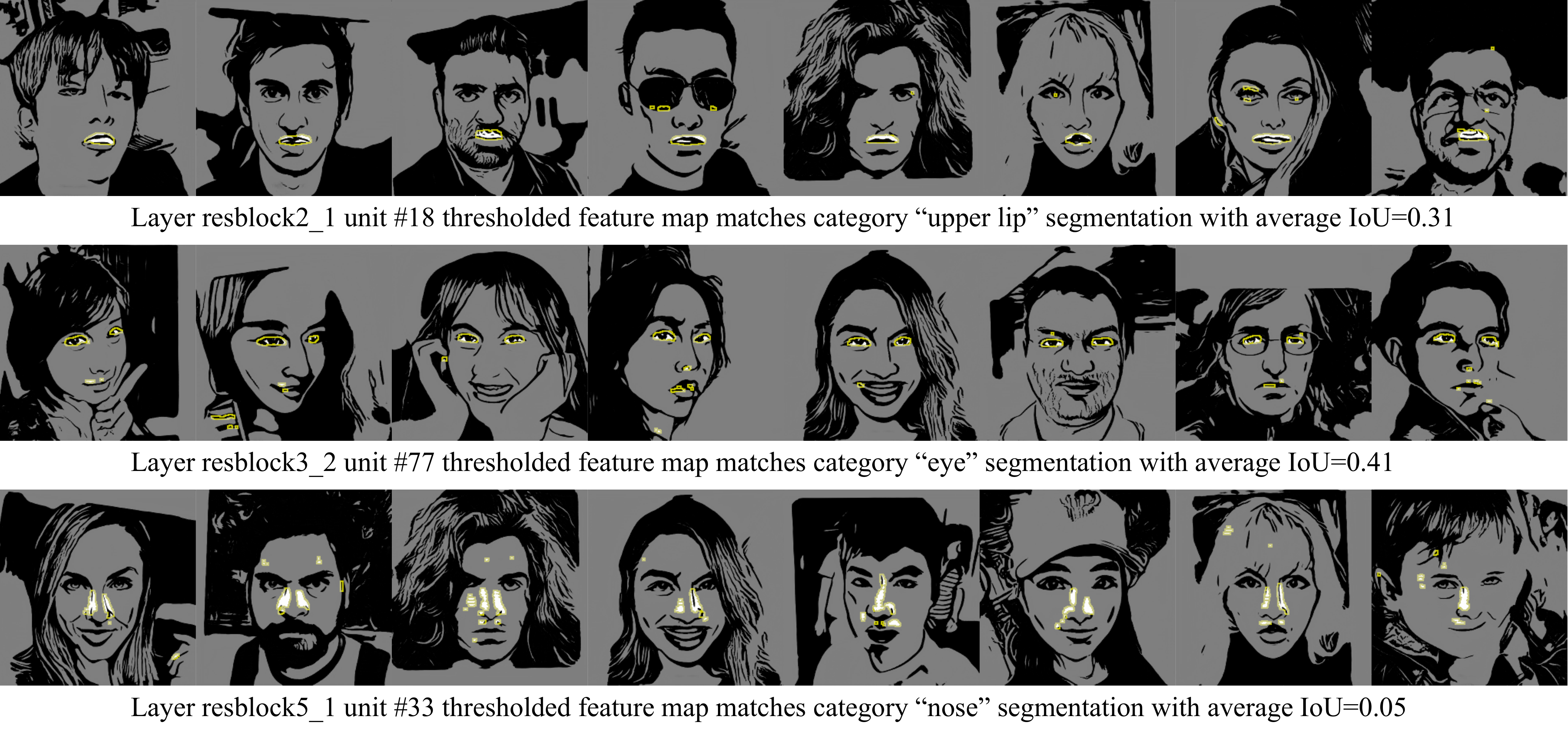}
\vspace{-0.15in}
\caption{\exten{Visualizing interpretable units. 
The rows from top to bottom show units that best match ``upper lip", ``eye", and ``nose" categories, respectively. 
The IoU is measured over the full test set of 154 images. 
For each unit, eight images with top IoU are shown, and the masks of thresholding the upsampled feature map ($F_{u}^{\uparrow} > t_{u,r}$) are outlined in yellow.}}
\label{fig:dissect}
\vspace{-0.1in}
\end{figure*}

\exten{To model the style similarity, we explored existing style losses~\cite{JingYFYS17} including Gram-matrix-based loss~\cite{GatysEB16} and histogram-based loss~\cite{WilmotRB17}.
We found histogram loss is better for measuring line drawing style differences. Given the generated \lyj{APDrawing} $A$ and the target style \lyj{APDrawing} $B$, histogram matching is performed to match feature activations of $A$ to feature activations of $B$. 
We use VGG-19~\cite{SimonyanZ14a} to extract features and take five feature activations for style representation (`conv1\_1', `conv2\_1', `conv3\_1', `conv4\_1', `conv5\_1').
Denote the $i$-th feature activation as $O_i$, and the $j$-th channel in $O_i$ as $O_{ij}$,
we compute the normalized histogram for $O_{ij}$ of $A$ and match it to the normalized histogram for $O_{ij}$ of $B$, thus obtaining the remapped activations. The process is repeated for each channel in $O_i$.
Denote the $i$-th \YL{remapped} activations as $R_i(O_i,A,B)$.
The histogram loss is defined as:
\begin{equation}
L_{histogram}(A,B) = \sum_{i=1}^{5} ||O_i(A) - R_i(O_i,A,B)||
\end{equation}
Then we set $L_{style}$ in Eq.~(\ref{eq:vec_op}) to $L_{histogram}$. We randomly initialize the style feature vector and use an Adam optimizer with learning rate $0.05$ to optimize the vector.
Some examples of the optimization process and results are shown in Figs.~\ref{fig:style_optim}c \rev{and \ref{fig:style_optim}d}.
}

\section{\exten{Generator Dissection}}

\lyj{Our model can successfully learn to generate good looking APDrawings in multiple styles using a single network, and can generate APDrawings of ``new styles'' unseen in the training data. To better interpret our model,}
\exten{we explore the semantic meaning of convolution layers in the APDrawing generator $G$ by visualizing feature maps and analyzing their relation to face semantics.
Following~\cite{BauZSZTFT19}, we measure the spatial agreement between thresholded feature map and facial part segmentation with intersection-over-union (IoU).
For a convolution layer unit $u$ and a
facial part region $r$ (e.g., upper lip, left eye, etc.), \lyj{denote the feature map of $u$ as $F_u$, the upsampled feature map as $F_{u}^{\uparrow}$, and the facial part region $r$'s segmentation\footnote{\lyj{A face parsing network~\cite{GuBY0WY19} is used for facial part segmentation.}} as $S_r$. The IoU is calculated as}
\begin{equation}
\begin{aligned}
IoU_{u,r} &= \frac{\mathbb{E}_{p} |(F_{u}^{\uparrow} > t_{u,r}) \cap S_r(p)|}{\mathbb{E}_{p} |(F_{u}^{\uparrow} > t_{u,r}) \cup S_r(p)|}\\
t_{u,r} &= \arg \max_{t} \frac{I(F_{u}^{\uparrow} > t; S_r(p))}{H(F_{u}^{\uparrow} > t; S_r(p))}
\label{eq:dissect}
\end{aligned}
\end{equation}
where $p$ denotes face photos sampled from the face photo domain $\mathcal{P}$, the IoU is measured over a test set of face photos. \lyj{The map $F_{u}^{\uparrow} > t_{u,r}$} produces a binary mask by thresholding the upsampled feature map at a fixed level $t_{u,r}$.
$S_r$ is a binary mask in which the foreground contains pixels \lyj{belonging} to the facial part region $r$.
The threshold $t_{u,r}$ is computed by maximizing the information quality $I/H$ where $H$ is the joint entropy and $I$ is the mutual information~\cite{Wijaya2017}.
}

\exten{We use $IoU_{u,r}$ to find the facial parts related to each convolution layer unit and label each unit with the facial part that best matches it. 
The units with $\max_{r} IoU_{u,r} > 0.05$ are called ``interpretable'' units.
Figure~\ref{fig:dissect} shows some examples of interpretable units with different labels.
\lyj{Among 5,505 convolution units in our generator $G$, 594 of them are interpretable units, showing that face semantic information is learned and incorporated during APDrawing generation.}}

\section{Experiments}

We implemented our \lyj{model} in PyTorch. All experiments are performed on a computer with a Titan Xp GPU.
The parameters in Eq.~(\ref{eq:loss}) are $\lambda_1=5-\rev{\frac{4.5i}{N}}$, $\lambda_2=5$, $\lambda_3=\rev{\frac{4.5i}{N}}$, $\lambda_4=1$, \lyjrev{$\lambda_5=0.5 \cdot {\bf 1}_{\{i > 100\}}(i)$, where ${\bf 1}_A$ is the indicator function,} $i$ is the current epoch number, and \rev{$N$} is the total epoch number \rev{($N=300$)}.
\rev{We apply the quality loss after 100 epochs \lyjrev{so that the model can learn a proper drawing first and is then optimized towards better quality.}}

\subsection{Experiment Setup}
\label{subsec:setup}

{\bf Data.} We collect face photos and APDrawings from the Internet and construct \lyj{(1) a training corpus of 798 face photos and 625 delicate portrait line drawings, and (2) a test set of 154 face photos.}
Among the collected drawings, \lyj{(1) 84 are labeled with artist {\it Charles Burns}, 48 are labeled with artist {\it Yann Legendre}, 88 are labeled with artist {\it Kathryn Rathke}, and 212 are from the website {\it vectorportral.com}, and (2) others have no tagged author/source information.}
We observed that both Charles Burns and Yann Legendre use similar \rev{thin parallel lines} to draw shadows, and so we merged drawings of these two artists into style1. We \lyj{choose} the drawings of Kathryn Rathke as style2 and the drawings of vectorportral as style3. \lyj{Styles 1 and 2} have distinctive features:  Kathryn Rathke uses flowing lines but few dark regions, \lyj{while} vectorportral uses thick lines and large dark regions.
All the training images are resized and cropped to $512 \times 512$ pixels.

\begin{figure}[t]
\centering
\includegraphics[width = 0.49\textwidth]{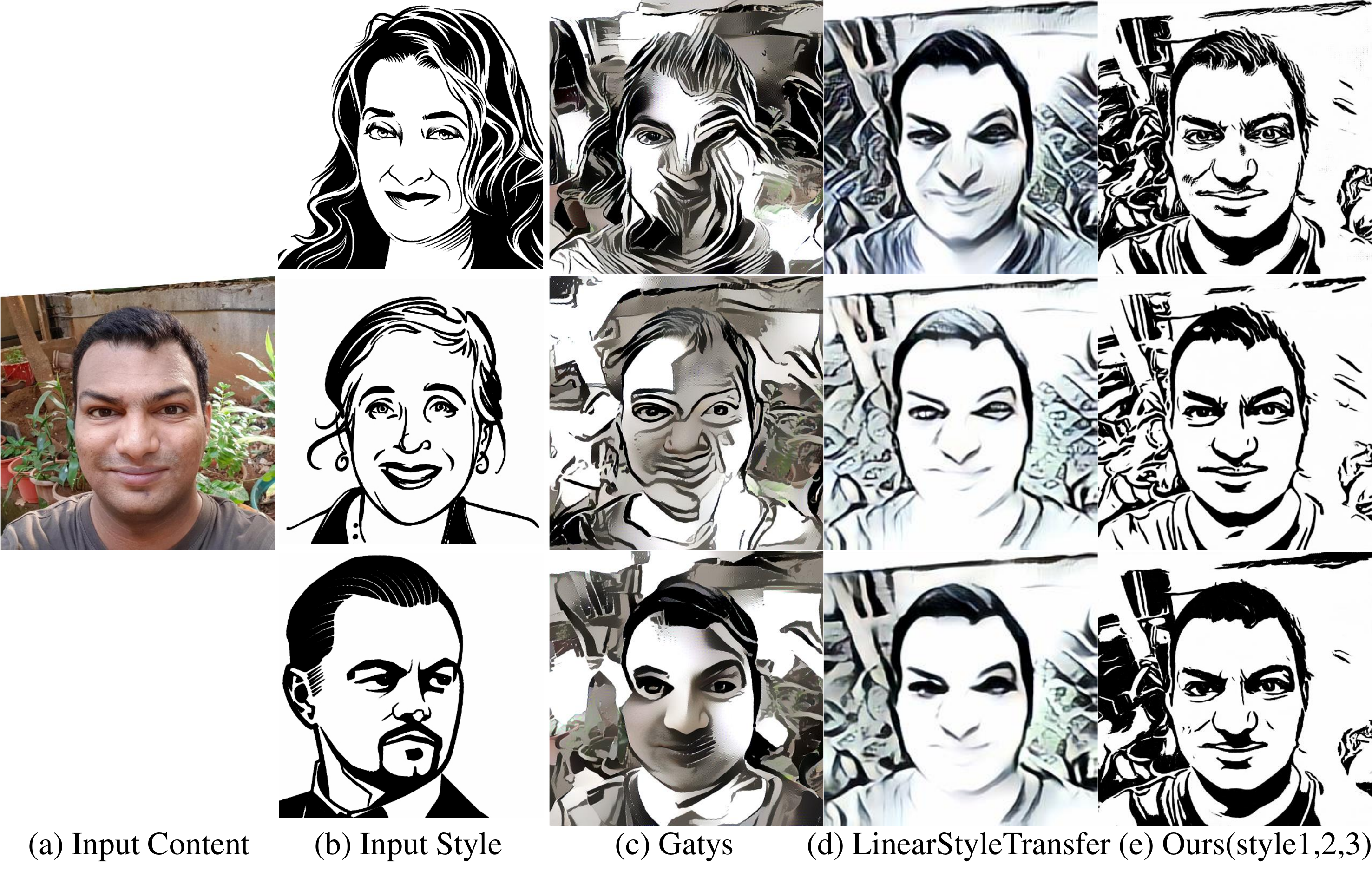}
\vspace{-0.15in}
\caption{Comparison with two state-of-the-art neural style transfer methods, i.e., Gatys~\cite{GatysEB16} and LinearStyleTransfer~\cite{LiLK019}. }
\vspace{-0.05in}
\label{fig:compare0}
\end{figure}

\begin{figure*}[thb]
\centering
\includegraphics[width = 0.7\textwidth]{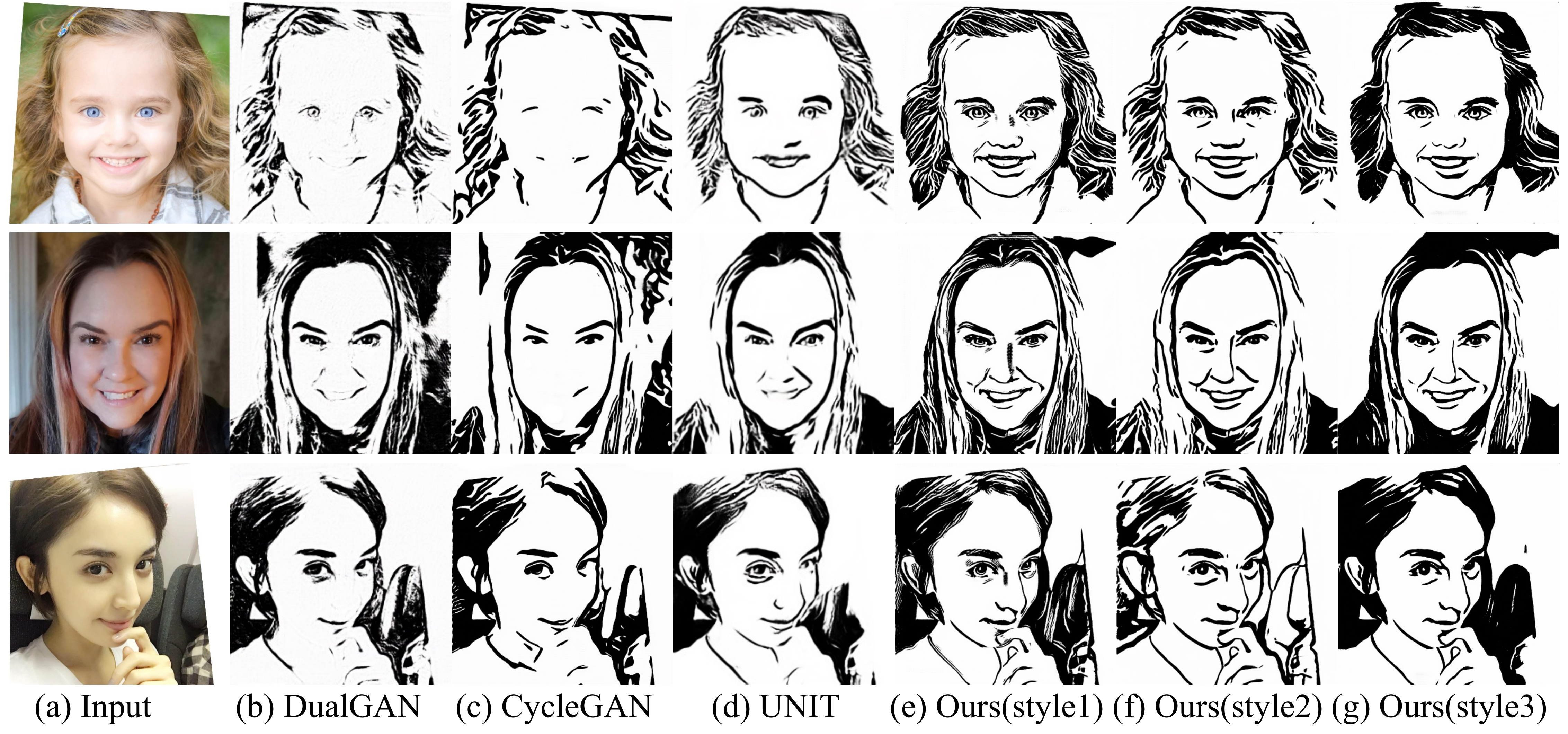}
\vspace{-0.2in}
\caption{\rev{Comparison with three single-modal unpaired image-to-image translation methods: DualGAN~\cite{YiZTG17}, CycleGAN~\cite{ZhuPIE17}, UNIT~\cite{LiuBK17}. \lyjrev{All methods are trained using the same training corpus with both real and synthesized drawings.}}}
\label{fig:compare1}
\end{figure*}

\begin{figure*}[thb]
\centering
\includegraphics[width = 1.0\textwidth]{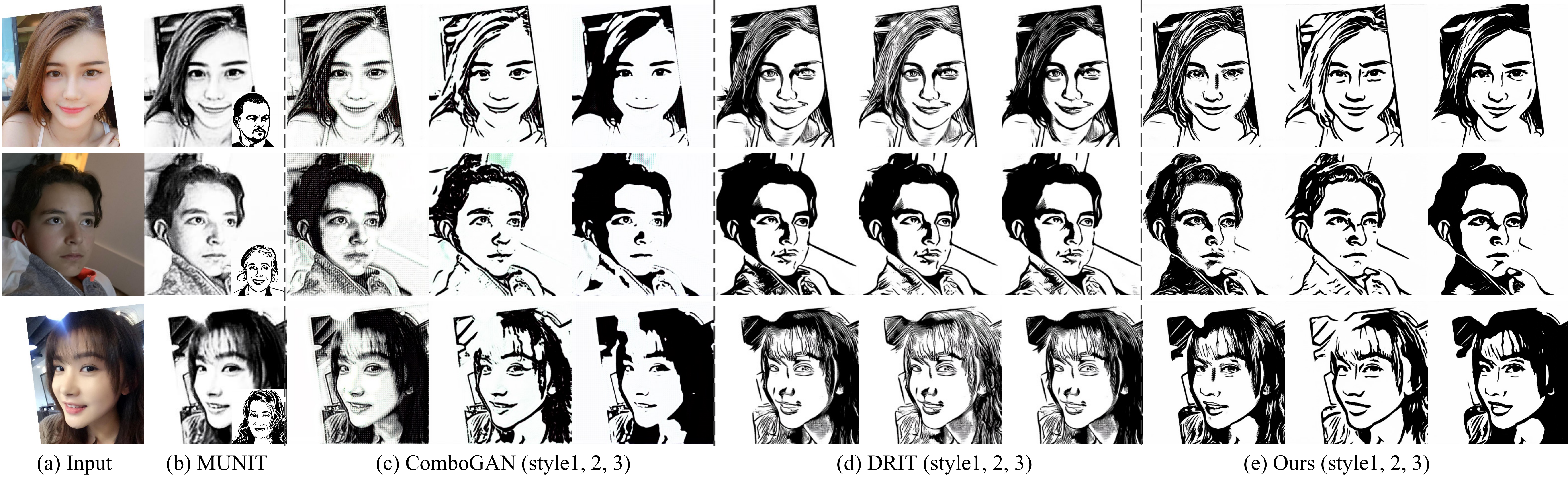}
\vspace{-0.2in}
\caption{\rev{Comparison with three unpaired image-to-image translation methods that can deal with multi-modal or multi-domain translation: MUNIT~\cite{HuangLBK18}, ComboGAN~\cite{AnooshehATG18}, DRIT~\cite{LeeTHSY18}. \lyjrev{All methods are trained using the same training corpus with both real and synthesized drawings.}}}
\label{fig:compare2}
\end{figure*}

\lyj{{\bf Training process.} It includes two steps: 
(1) training classifier $C$ and (2) Training  our  model.} We first train a style classifier $C$ (Section~\ref{sssec:archg}) with the tagged drawings and data augmentation (including random rotation, translation and scaling).
To balance the number of drawings in each style, \lyj{we take all drawings from the first and second styles, but only part of the third style in the training stage of $C$, in order to achieve balanced training for different styles.}
\lyj{In the second step,} we use the trained classifier to obtain \lyj{style feature vectors} for all 625 drawings.
We further augment training data using synthesized drawings.
Training our network with the mixed data of \lyj{real and synthesized drawings} results in high-quality generation for all three styles; see Figs.~\ref{fig:compare0}-\ref{fig:compare2} for some examples, where our results of styles 1, 2, 3 are generated by \lyj{feeding in style feature vectors $[1,0,0], [0,1,0], [0,0,1]$, respectively.}

\subsection{Comparisons}

We compare our method with two state-of-the-art neural style transfer methods: Gatys~\cite{GatysEB16}, LinearStyleTransfer~\cite{LiLK019}, and \lyj{six} unpaired image-to-image translation methods:  DualGAN~\cite{YiZTG17}, CycleGAN~\cite{ZhuPIE17}, UNIT~\cite{LiuBK17}, MUNIT~\cite{HuangLBK18}, ComboGAN~\cite{AnooshehATG18} \exten{and DRIT~\cite{LeeTHSY18}}.

\rev{For the two neural style transfer methods, i.e., Gatys~\cite{GatysEB16} and LinearStyleTransfer~\cite{LiLK019}, \lyjrev{their} inputs are a content image (face photo) and a style image (one of the collected artist line drawings). 
For the six unpaired image-to-image translation methods, i.e., DualGAN~\cite{YiZTG17}, CycleGAN~\cite{ZhuPIE17}, UNIT~\cite{LiuBK17}, MUNIT~\cite{HuangLBK18}, ComboGAN~\cite{AnooshehATG18} and DRIT~\cite{LeeTHSY18}, we retrained each comparison model using our training set, which consists of 978 photos, and both 625 collected real drawings and 353 synthesized drawings.}

Comparisons with neural style transfer methods are shown in Fig.~\ref{fig:compare0}. Gatys' method fails to capture \lyj{APDrawing} styles because it uses the Gram matrix to model style as texture, \lyj{but APDrawings have little texture. LinearStyleTransfer produces visually better results, although they are still not desired APDrawings: the generated drawings have many thick lines and they are produced in a rough manner.} Compared to these example-guided style transfer methods, our method learns from a set of APDrawings and generates delicate results for all three styles.

Comparisons with single-modal unpaired image-to-image translation methods are shown in Fig.~\ref{fig:compare1}. DualGAN and CycleGAN are both based on strict cycle-consistency loss. \lyj{This causes a dilemma in photo-to-APDrawing translation: either a generated drawing looks like a real drawing (i.e., close to binary, containing large uniform regions) which cannot properly reconstruct the original photo, or a generated drawing has good reconstruction with grayscale changes but which does not look like a real drawing. Meanwhile,}
compared to CycleGAN, DualGAN is more grayscale-like, less abstract and worse in line drawing style.
UNIT adopts feature-level cycle-consistency loss, \lyj{which less constrains the results  at the image level,} making the face appear deformed.
In comparison, our results both preserve face structure and have good image and line quality.

Comparisons with unpaired image-to-image translation methods that can deal with multi-modal or multi-domain translation are shown in Fig.~\ref{fig:compare2}.
Results show that MUNIT does not capture \lyj{the APDrawing styles} and the results are more similar to a pencil drawing with shading and many gray regions.
ComboGAN fails to capture all three representative styles, which performs slightly better on styles 2 and 3 than style 1.
\exten{DRIT also fails to capture all three representative styles; the results of all three styles look similar and only approximate the target styles 1 and 3. In comparison,}
our method generates distinctive results for three styles and reproduces the characteristics for each style well.

\subsection{\exten{Quantitative Evaluation}}

\begin{table}[t]
  \centering
  \caption{\rev{User study results. The $i$-th row shows the percentages of different methods (ComboGAN~\cite{AnooshehATG18}, CycleGAN~\cite{ZhuPIE17}, our conference version (Ours-pre)~\cite{YiLLR20} and Ours) being ranked as the $i$-th among four methods.}}
  {
  \begin{tabular}{c|c|c|c|c}
  \hline
     & ComboGAN & CycleGAN & Ours-pre & Ours\\
  \hline
  Rank1 & 23.7\%  &  8.1\%  &  23.1\%  &  \textbf{45.1\%}\\
  \hline
  Rank2 & 12.2\%  &  8.9\%  &  46.4\%  &  32.6\%\\
  \hline
  Rank3 & 35.1\%  &  27.9\%  &  21.9\%  &  15.1\%\\
  \hline
  Rank4 & 29.0\%  &  55.2\%  &  8.6\%  &  7.2\%\\
  \hline
  \end{tabular}}
  \label{table:userstudy}
\end{table}

\begin{table}[t]
  \centering
  \caption{\rev{Analysis of variance (ANOVA) results for pairwise comparisons.}}
  \begin{tabular}{c|c}
  \hline
  Pairwise comparison & Ranked Best\\
  \hline
  Ours vs. ComboGAN & \rev{$p$=3.57e-7}\\
  Ours vs. CycleGAN & \rev{$p$=1.32e-22}\\
  Ours vs. Ours-pre & \rev{$p$=2.09e-11}\\
  \hline
  \end{tabular}
  \label{table:anova}
\end{table}

\begin{figure}[t]
\centering
\includegraphics[width=0.96\columnwidth]{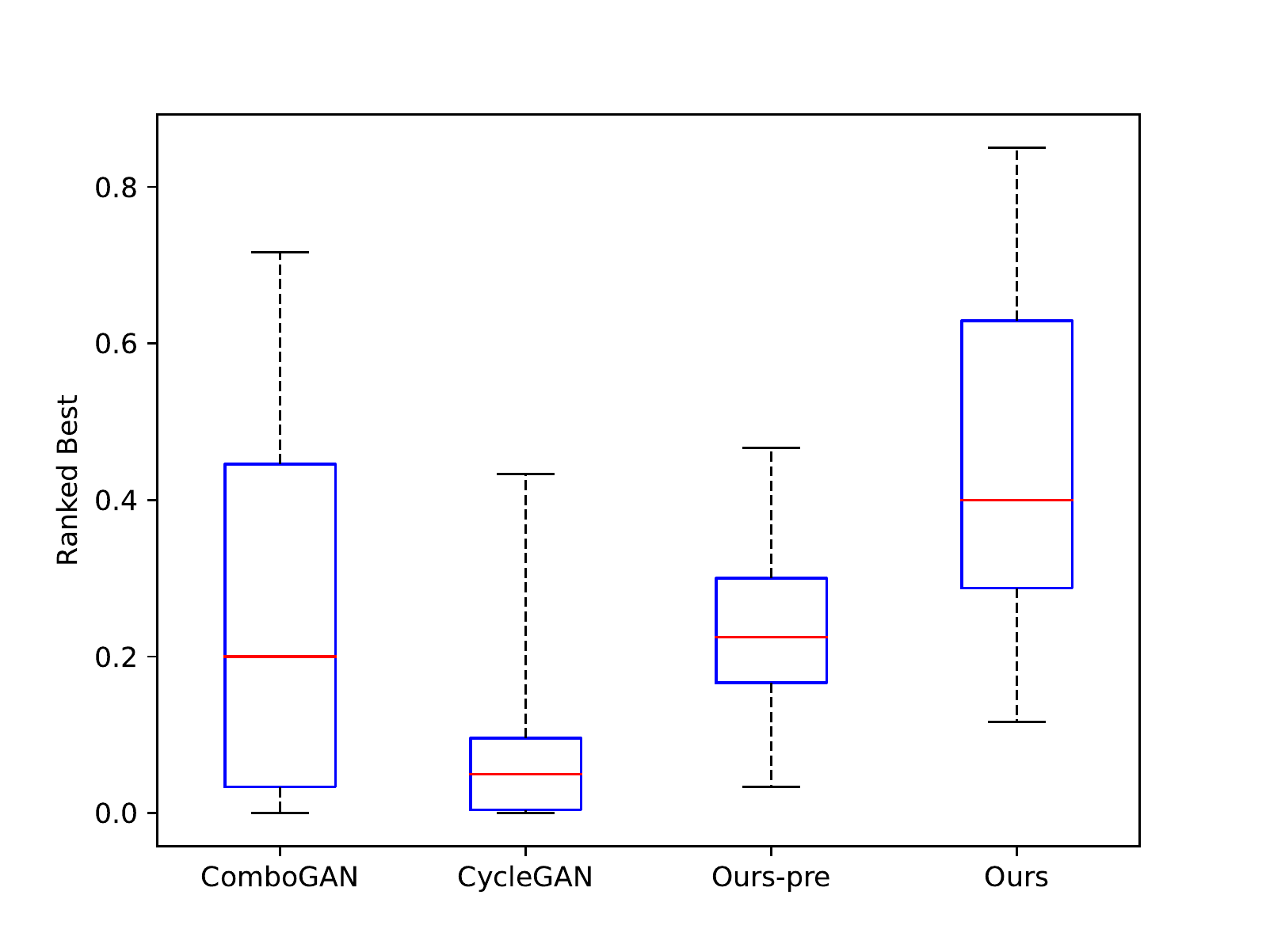}
\vspace{-0.20in}
\caption{\rev{Test boxplot \cite{Hogg1987} of four methods. In each box, the central red line indicates the median, and the bottom and top blue edges of the box indicate the 25\% and 75\% percentiles respectively. The dashed black line extends to the extreme data points.}}
\label{fig:boxplot}
\vspace{-0.10in}
\end{figure}

\exten{{\bf User study.}
\lyj{Considering the artistic merits of portrait line drawings, we conduct a user study to compare our method with CycleGAN \cite{ZhuPIE17}, ComboGAN \cite{AnooshehATG18} and our conference version (Ours-pre) \cite{YiLLR20}. LinearStyleTransfer, Gatys, DualGAN and UNIT are not included because of their lower visual quality. MUNIT and DRIT are not included because they obviously do not capture the target styles.}
We randomly sample 60 groups of images/drawings from the test set: 20 for style1 comparison, 20 for style2 and 20 for style3.
Before the test, the participants went through some practice examples, and were given guidelines about the standard of good portrait line drawings.
During the test, \lyj{participants (1) were shown a photo, a real drawing (the style reference) and 4 generated drawings at a time, and (2) were asked to sort 4 results from best to worst.}
\rev{54 participants} attended the user study and \rev{3,240 votes} were collected in total.}

\begin{table}[t]
  \centering
  \caption{\rev{Fr\'{e}chet Inception Distance (FID) of our method and four multi-modal image translation methods. The FID values are computed between the set of generated APDrawings of each style and the collected true drawings of the corresponding style.}}
  \begin{tabular}{c|c|c|c}
  \hline
  Methods & Style1 \rev{$\downarrow$} & Style2 \rev{$\downarrow$} & Style3 \rev{$\downarrow$}\\
  \hline
  MUNIT~\cite{HuangLBK18} & \rev{194.3} & \rev{267.4} & \rev{242.8}\\
  ComboGAN~\cite{AnooshehATG18} & \rev{184.9} & \rev{144.1} & \rev{141.4}\\
  \exten{DRIT~\cite{LeeTHSY18}} & \rev{82.8} & \rev{135.0} & \rev{119.9}\\
  \exten{Ours-pre~\cite{YiLLR20}} & 88.3 & 139.0 & 108.2\\
  \exten{Ours} & \textbf{81.2} & \textbf{114.3} & \textbf{89.7}\\
  \hline
  \end{tabular}
  \label{table:fid}
\end{table}

\begin{table}[t]
  \centering
  \caption{\rev{The scores predicted by quality metric model $M$ on the results of different methods. The score for each method is averaged on the test set. Higher quality score indicates better quality.}}
  \setlength\tabcolsep{2.0pt}
  \begin{tabular}{c|c|c|c|c|c}
  \hline
  Methods & Gatys & LST & DualGAN & CycleGAN & UNIT \\
  \hline
  Quality score & 0.37 & 0.35 & \rev{0.35} & \rev{0.35} & \rev{0.36} \\
  \hline
  Methods & MUNIT & ComboGAN & DRIT & Ours-pre & Ours\\
  \hline
  Quality score & \rev{0.33} & \rev{0.40} & \rev{0.44} & 0.45 & \textbf{0.51}\\
  \hline
  \end{tabular}
  \label{table:metric_model}
\end{table}

\exten{\lyjrev{Results of the percentages of each method ranked as 1st, 2nd, 3rd and 4th are summarized in Table~\ref{table:userstudy}.
Our method ranks the best with \rev{$45.1\%$} of the votes, which is higher than the other methods, i.e., ComboGAN, CycleGAN and our conference version, which rank the best in \rev{$23.7\%$, $8.1\%$ and $23.1\%$} of the \lyj{votes}.
The average rank of our method is \rev{1.84}, lower compared to CycleGAN's \rev{3.30}, ComboGAN's \rev{2.69} and our conference version's \rev{2.16}.}
We then conduct analysis of variance (ANOVA) between our method and \lyj{each of other methods on the percentage of being ranked best \YL{by individual users}.}
Pairwise ANOVA results are shown in Table \ref{table:anova}.
All of the $p$-values are $\ll 0.01$, justifying that the rejection of the null hypothesis and the differences between the means of our method and \lyj{each of other \YL{three} methods} (ComboGAN, CycleGAN or our conference version) are statistically significant.
A test boxplot of four methods is shown in \minorrev{Fig.}~\ref{fig:boxplot}.
These results demonstrate that our method outperforms other methods.
All generated drawings evaluated in the user study are presented in the appendix.
}

\begin{figure}[t]
\centering
\includegraphics[width = 0.5\textwidth]{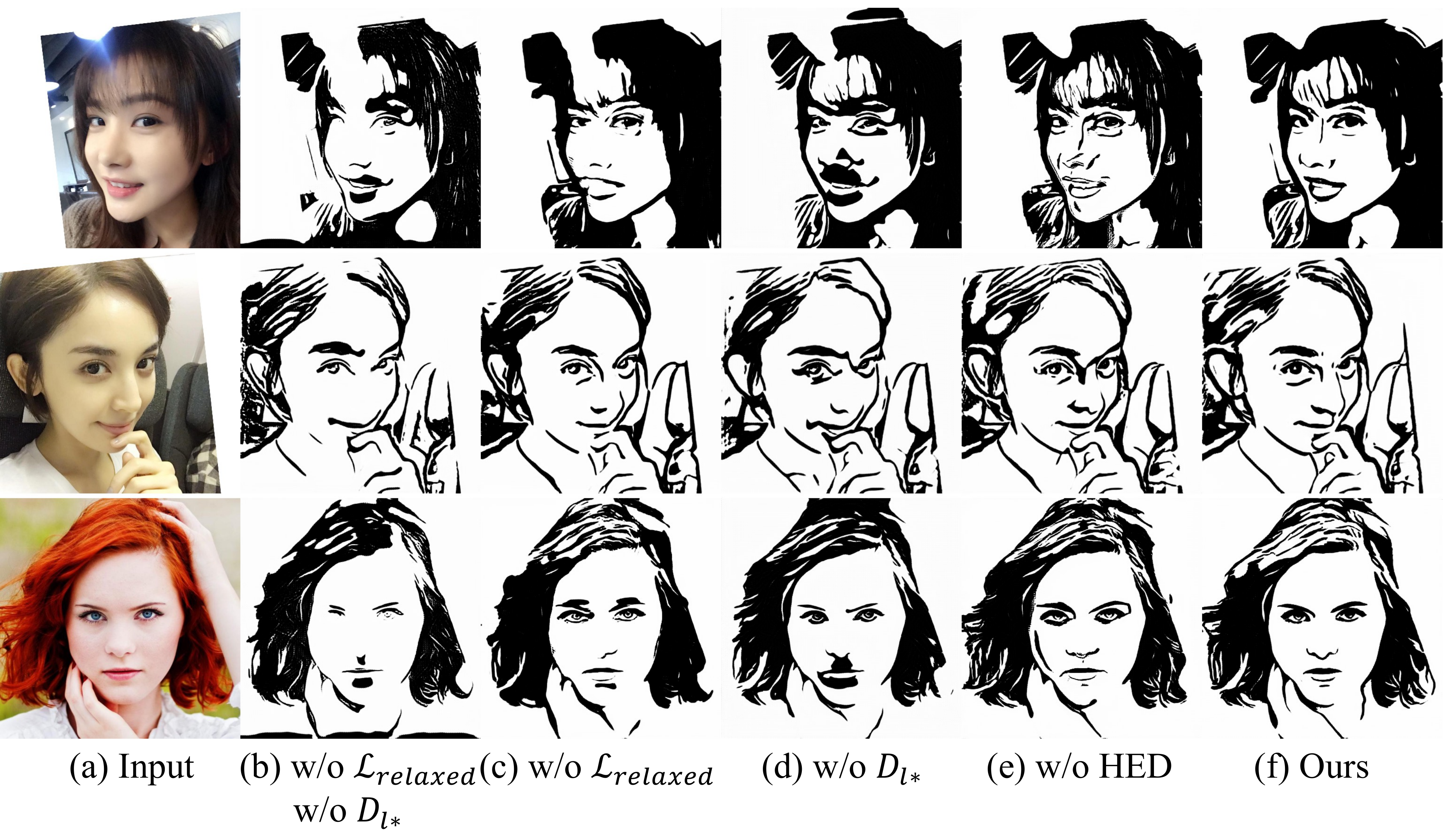}
\vspace{-0.2in}
\caption{\rev{Ablation study: (a) input photos, (b) results of removing relaxed cycle-consistency loss (i.e. using $L_1$ loss) and removing local discriminators, (c) results of removing relaxed cycle-consistency loss, (d) results of removing local discriminators, (e) results of removing HED in calculating relaxed cycle-consistency loss, (f) our results.}}
\label{fig:ablation}
\vspace{-0.1in}
\end{figure}

{\bf GAN Metric Evaluation.}
We adopt the Fr\'{e}chet Inception Distance (FID)~\cite{heusel2017gans} to evaluate the similarity between the distributions of two drawing sets --- one is the set of generated APDrawings for one style and the other is the set of collected true drawings for this style --- where lower FID indicates better similarity.
\lyj{By changing the input style feature vector, we transform all face photos in the test set into three styles of APDrawings.}
The FID values between the set of generated APDrawings of each style and the collected drawings of the corresponding style are computed and summarized in Table~\ref{table:fid}.
The results show that compared with the other multi-modal generation methods (MUNIT~\cite{HuangLBK18}, ComboGAN~\cite{AnooshehATG18}, \exten{DRIT~\cite{LeeTHSY18} and our conference version~\cite{YiLLR20}}), our method has lower FID on all three styles, indicating our method generates a closer distribution to the distribution of true drawings.

\exten{{\bf Quality metric model evaluation.}
We apply the trained quality metric model $M$ on generated drawings of different methods, and the quality scores are listed in Table~\ref{table:metric_model}.  The score for each method is averaged on the test set. Our method achieves the highest score, \lyj{indicating that our generated results have the best perceptual quality according to the trained metric model.}
}

\lyj{{\bf More test results.}
In addition to photos collected from Internet, we also test our method on photos from \minorrev{the} CelebAMask-HQ Dataset~\cite{abs-1907-11922}. The results are summarized in Appendix E.3.}

\begin{figure*}[t]
\begin{center}
\includegraphics[width=1.0\textwidth]{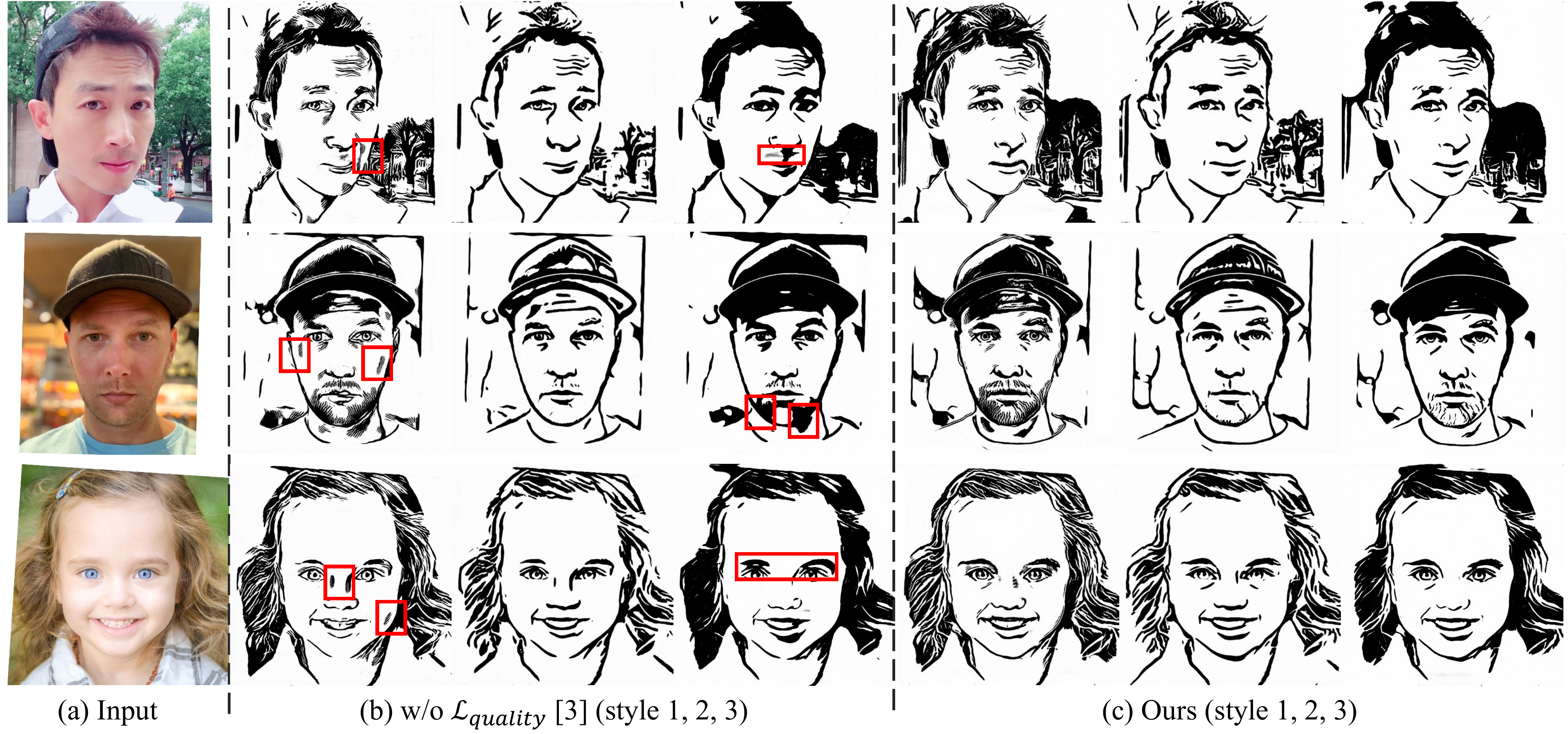}
\end{center}
\vspace{-0.22in}
\caption{\exten{Ablation study on quality loss: (a) input photos, (b) results of removing quality loss, (c) our results. \lyj{Artifacts are highlighted in red boxes.}}}
\label{fig:ablation_quality}
\vspace{-0.12in}
\end{figure*}

\subsection{Ablation Study}

\begin{table}[t]
  \centering
  \caption{\rev{Fr\'{e}chet Inception Distance (FID) of the ablation studies. The FID values are computed between the set of generated APDrawings of each style and the collected true drawings of the corresponding style.}}
  {
  \begin{tabular}{c|ccc|c}
  \hline
  Methods & Style1 $\downarrow$ & Style2 $\downarrow$ & Style3 $\downarrow$ & Avg $\Delta$\\
  \hline
  w/o $L_{relaxed}$, w/o $D_{l*}$ & 102.9 & 126.8 & 105.1 & 16.53\\
  w/o $L_{relaxed}$ & 88.0 & 132.9 & 107.0 & 14.23 \\
  w/o $D_{l*}$ & 89.4 & 142.3 & 101.4 & 15.97\\
  w/o HED & 84.1 & 114.8 & 103.4 & 5.70\\
  w/o $L_{quality}$~\cite{YiLLR20} & 88.3 & 139.0 & 108.2 & 16.77\\
  Ours & \textbf{81.2} & \textbf{114.3} & \textbf{89.7} & /\\
  \hline
  \end{tabular}}
  \label{table:fid2}
\end{table}

\lyj{We perform an ablation study based on \minorrev{the} four key ingredients of our model: (1) relaxed cycle consistency loss, (2) the quality loss based on the quality metric model, (3) local discriminators, and (4) HED edge extraction.
Results show that they are all essential to our model. In Appendix D, three more ablation studies are presented: the first focuses on the style feature vector and style loss, the second focuses on the truncation loss, and the third focuses on how face region information is utilized in the discriminator.}

As shown in Fig.~\ref{fig:ablation}b, without relaxed cycle consistency loss and local discriminators, facial features are often missing, \rev{e.g., the nose is missing in all three rows, and eye details are missing in the first and third rows}.
Removing only relaxed cycle consistency loss (Fig.~\ref{fig:ablation}c) preserves more facial feature regions \rev{(e.g., the nose in the second row)} when compared to Fig.~\ref{fig:ablation}b, but some parts \lyjrev{(e.g., the nose in the third row)} are still missing \lyjrev{compared to our method (Fig.~\ref{fig:ablation}f)}.
Removing only local discriminators (Fig.~\ref{fig:ablation}d) produces few missing parts: although the results are much better than Fig.~\ref{fig:ablation}b in terms of preserving facial structure, some facial features are not drawn in the desired manner, i.e., some black regions or shadows (that are usually drawn near facial boundaries or hair) appear near the nose.
When both relaxed cycle consistency loss and local discriminators are used, results (Fig.~\ref{fig:ablation}f) preserve all facial feature regions and no undesired black regions or shadows appear in faces.
\rev{These results show that both relaxed cycle consistency loss and local discriminators help to preserve facial feature regions and are complementary to each other
: (1) the relaxed cycle loss works in a more global and general way, it alleviates the need to hide information and helps preserve outlines (since lines are more easily missing in nose, eyes \lyjrev{and lips regions, their effects} on these regions are more visible), \lyjrev{and} (2) \lyjrev{as a comparison,} the local discriminators work in a local way, dedicated to eyes, nose and lips, improving drawings and eliminating artifacts in these local regions.}

As shown in Fig.~\ref{fig:ablation}e, without HED edge extraction in the relaxed cycle consistency loss calculation \rev{(i.e., calculating LPIPS perceptual similarity between the input and reconstructed photo)}, the lines are often discontinuous or blurred, \rev{e.g., the nose outlines in the second row are discontinuous (upper right), and the noses in the first and third rows are blurred \lyjrev{and} messy}. 
In comparison, our results have clear, sharp and continuous lines, \lyj{demonstrating} that using HED edge extraction helps the model to generate clearer and more complete lines.

\exten{
As shown in Fig.~\ref{fig:ablation_quality}(b), without the quality loss, the results contain more artifacts including undesired dark regions and parallel lines on the face \lyj{(highlighted in red boxes)}.
In comparison, our results in Fig.~\ref{fig:ablation_quality}(c) are cleaner and of better quality.
}

\rev{The quantitative evaluation of the above ablation studies are reported in Table~\ref{table:fid2}. FID scores of our results are lower (better) in all three styles than the ablated versions.
We further compute the average difference between our FID and each ablated version, shown as ``Avg $\Delta$''.
}

\rev{\lyjrev{{\bf Contributions of each component.}} (1) Qualitative and quantitative results show that cycle consistency loss and local discriminators are complementary to each other, and work together to better preserve facial features:\lyjrev{ 
\begin{itemize}
    \item without both components, the average $\Delta$ is larger than without a single component;
    \item without a single component, the average $\Delta$ is also large, indicating these two components contribute largely to the final results.
\end{itemize}} 
\noindent (2) In addition, the visual differences between results of removing the quality loss and ours are easily visible, \lyjrev{and} the quantitative difference is also large, indicating the quality loss helps remove undesirable artifacts and improves quality.
(3) Compared to these three components, HED itself has smaller impact on the final results.}

\section{Conclusion}
In this paper, we propose a method for \lyj{high quality APDrawing} generation using asymmetric cycle mapping.
Our method can learn multi-style APDrawing generation from \lyjrev{web data of mixed styles} using an additional style feature \lyj{vector} input and a soft classification loss. 
\exten{In particular, our method \lyj{makes use of unpaired training data and improves upon \cite{YiLLR20} in the following four aspects:} (1) a novel quality metric for \lyj{APDrawings} is proposed; (2) based on the quality metric, a new quality loss \lyjrev{that is consistent with human perception} is introduced to guide the model toward better looking drawings; (3) a ``new'' style \lyj{APDrawing generation mechanism} is proposed; and (4) the model is dissected by visualizing feature maps and \lyj{exploring} face semantics.
}
Experiments and a user study demonstrate that our method can \lyj{(1) generate high quality and distinctive APDrawing results} for \exten{\lyjrev{the styles} in training data and new unseen styles,} and \lyj{(2)} outperforms state-of-the-art methods.

\section*{\lyjrev{Acknowledgement}}

\lyjrev{This work was supported by the Natural Science Foundation of China (61725204, 72192821), Tsinghua University Initiative Scientific Research Program and Shanghai Municipal Science and Technology Major Project (2021SHZDZX0102).}


%

\appendices

\section{Overview}
\exten{This appendix includes \lyj{the following material}}:
\begin{itemize}
\item \exten{detailed design of the network architecture (Section~\ref{sec:network_detail})};
\item more style examples in the training set (Section~\ref{sec:style_example});
\item three more ablation studies \rev{and their quantitative evaluation results} (Section~\ref{sec:ablation});
\item all evaluation material used in the user study in Section \lyj{7.4} of the main paper (Section~\ref{ssec:compare60});
\item comparison with \lyj{APDrawingGAN++} (Section~\ref{ssec:apdrawinggan});
\item more test results on other face dataset (Section~\ref{ssec:more_test}).
\end{itemize}

\section{\exten{Details of Network Architecture}}
\label{sec:network_detail}

\exten{In the main paper, we summarize the flowchart of the network architecture in \lyj{Figure 5} and introduce the architecture design principle in \lyj{Section 4.2}. Here we present the fine details of our proposed network architecture in Figure~\ref{fig:network_details}.
We denote the output channel as $c$, convolution kernel size as $k$, and stride in a convolution layer as $s$.
`Norm' means the instance normalization layer and `LReLU' means the leaky ReLU with $\alpha=0.2$.}

\section{More Style Examples in Training Set}
\label{sec:style_example}

\begin{figure}[t]
\centering
\includegraphics[width = .5\textwidth]{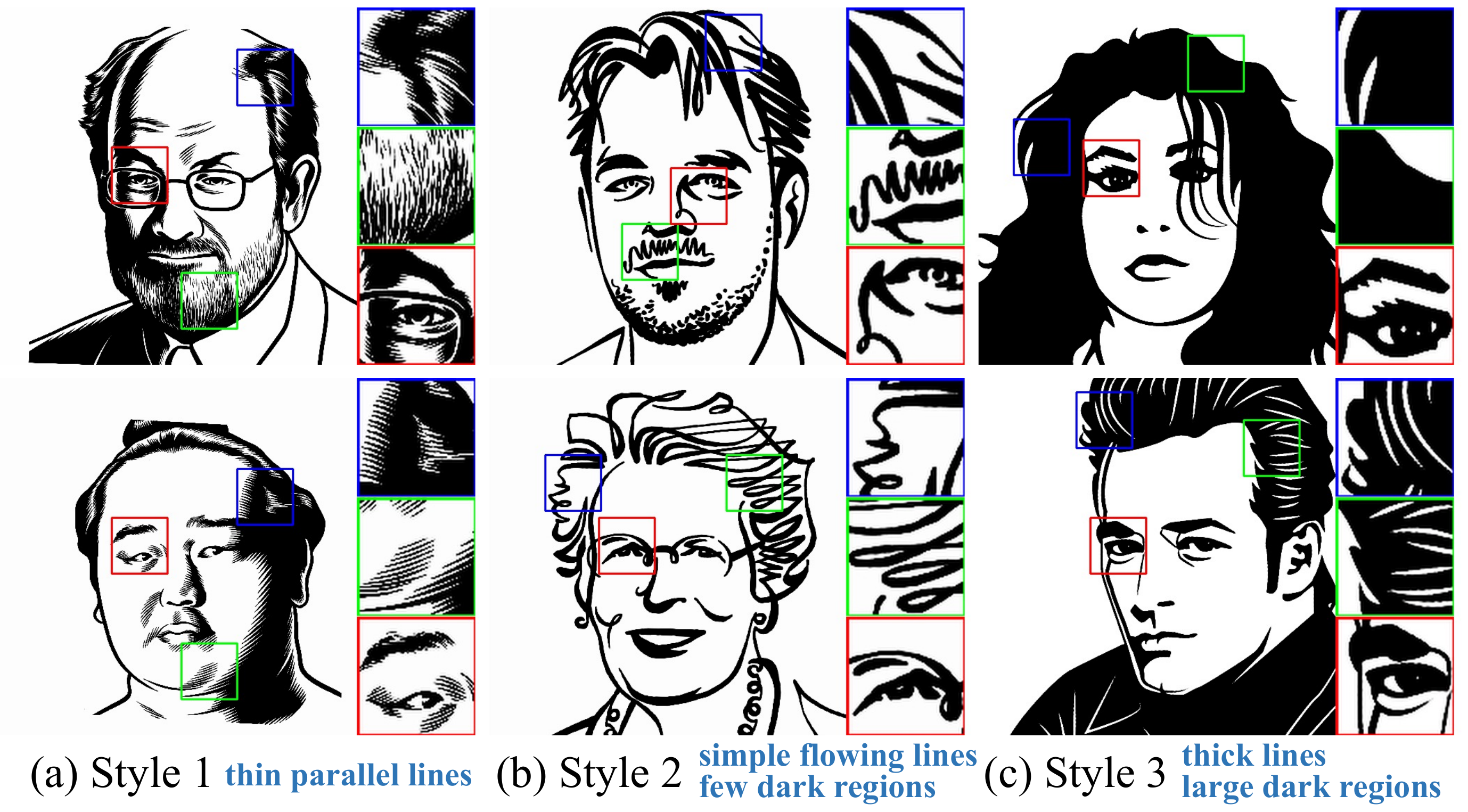}
\vspace{-0.2in}
\caption{\rev{More examples for three styles in the training set. Close-up views are shown alongside.}}
\label{fig:style_sample_more}
\vspace{-0.15in}
\end{figure}

In the main paper, we introduce the selected three representative styles from the collected data and show three examples in Figure 2: \lyj{(1)} the first style is from Yann Legendre and Charles Burns where \rev{thin} parallel lines are used to draw shadows; \lyj{(2)} the second style is from  Kathryn Rathke where few dark regions are used and facial features are drawn using simple flowing lines; \lyj{(3)} the third style is from {vectorportal.com} where continuous thick lines and large dark regions are utilized. Here we provide more examples in Figure~\ref{fig:style_sample_more}.

\section{Three More Ablation Studies}
\label{sec:ablation}

In Section \exten{7.4} of the main paper, we study \exten{some} key factors in our model, i.e., relaxed cycle-consistency loss, \lyj{quality loss,} local discriminators, and HED edge extraction.
Here, we present three more ablation studies: \lyj{(1) the first} focuses on the style feature and style loss, \lyj{(2) the second} focuses on the truncation loss, and \lyj{(3) the third} focuses on how face region information is utilized in the discriminator.

\begin{figure}[t]
\centering
\includegraphics[width = 0.48\textwidth]{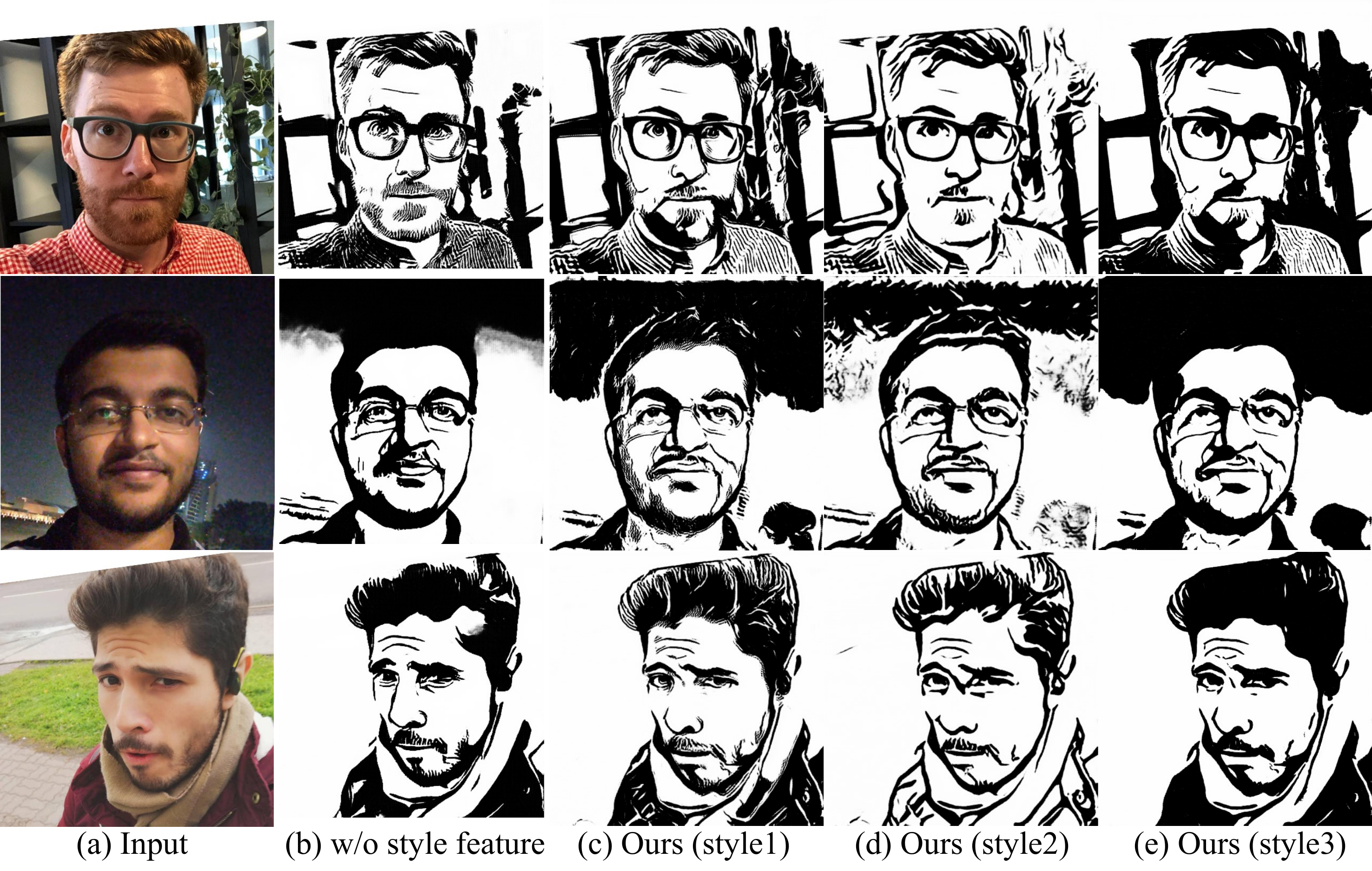}
\caption{\rev{Ablation study on style feature vector input and style loss. From left to right: input photos, results of removing style feature input and style loss, our results in styles 1, 2 and 3.}}
\label{fig:ablation_2}
\end{figure}

In our method, when inputting a face photo and a style feature \lyj{vector}, the system outputs an APDrawing with style specified by the style feature \lyj{vector}.
If we remove the style feature \lyj{vector} input and style loss from our system, when inputting a face photo, 
\rev{the model can output an APDrawing, but cannot generate APDrawings of different styles}.
Since the network is trained with mixed data, the output frequently exhibits different or mixed styles in different facial regions in an unpredictable way.
Three examples are illustrated in Figure \ref{fig:ablation_2}, in which all three photos contain a man face with beards.
On the top of Figure~\ref{fig:ablation_2}(b), the generated APDrawing shows a parallel line style in the beard \exten{and hair regions} \rev{(similar to style 1)}.
In the middle of Figure~\ref{fig:ablation_2}(b), thick line and dark region style appears \exten{near the eyes, hair and jawline regions} \rev{(similar to style 3)}.
At the bottom of Figure~\ref{fig:ablation_2}(b), the generated APDrawing shows mixed styles.
In comparison, as illustrated in Figures \ref{fig:ablation_2}(c-e), after introducing style feature \lyj{vector} and style loss, our method can generate APDrawing results for each distinctive style, specified by the input style feature \lyj{vector}.

\begin{figure}[t]
\centering
\includegraphics[width = 0.5\textwidth]{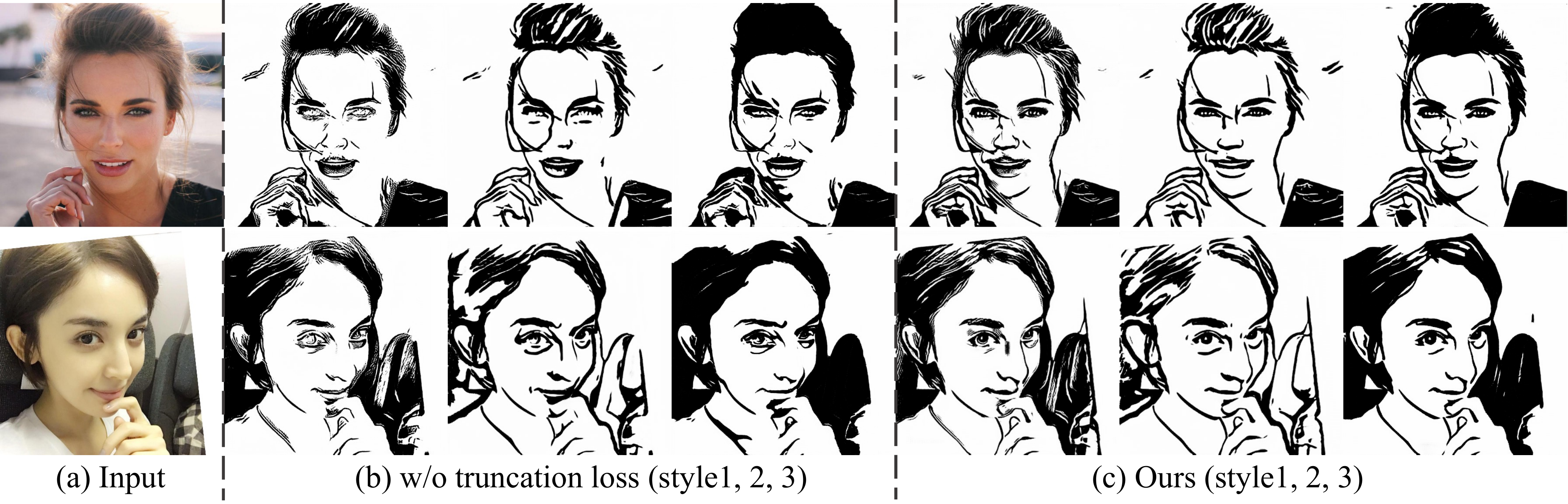}
\caption{\rev{Ablation study on truncation loss. From left to right: input photos, results of removing truncation loss (style1, 2, 3), and our results (style1, 2, 3).}}
\label{fig:ablation_3}
\end{figure}

We \lyj{also} study the role of truncation loss: \lyj{two examples are shown in Figure~\ref{fig:ablation_3}.}
The truncation loss is designed to prevent the generated drawings from hiding information in small values.
Without the truncation loss, the results sometimes do not draw full outlines of facial features (e.g., nose).
As shown in Figure~\ref{fig:ablation_3}b, 
\rev{the nose in the first row lacks the middle outline and the nose in the second row lacks the right outline.}
In comparison, by adding the truncation loss, our system can generate complete outlines of different facial features.

\begin{figure}[t]
\begin{center}
\includegraphics[width = 0.5\textwidth]{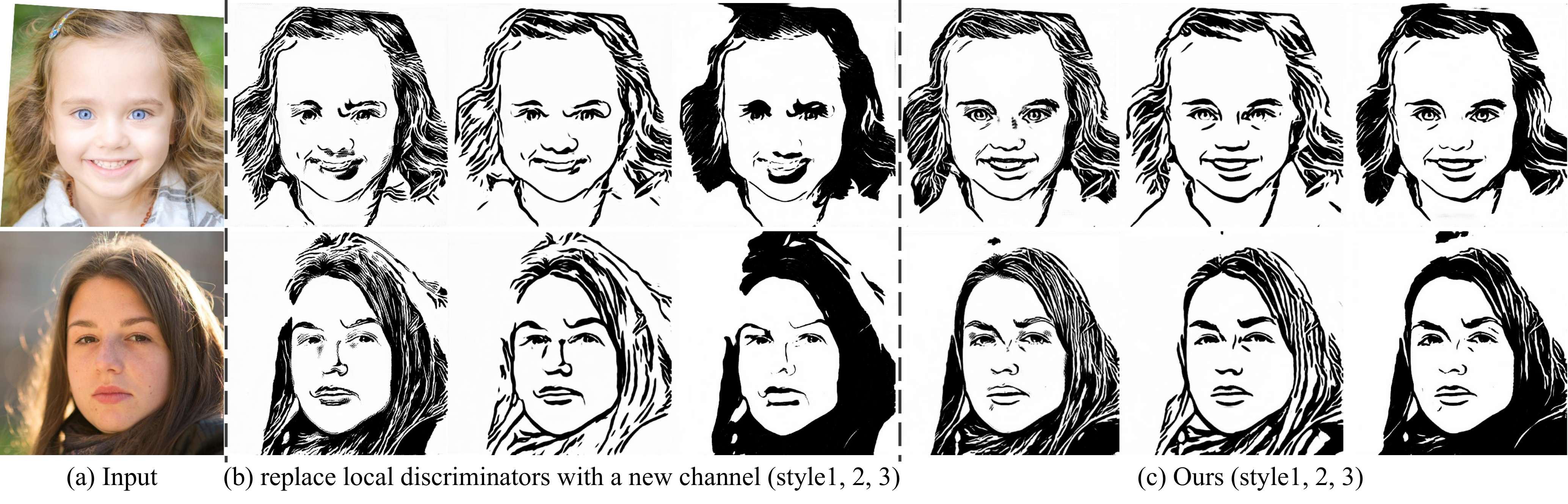}
\end{center}
\vspace{-0.15in}
\caption{
\rev{Comparison of results with our local discriminators (c) and replacing them with a new channel (b) for input photos (a).}
}
\label{fig:compare_addchannel}
\end{figure}

\begin{figure}[t]
\begin{center}
\includegraphics[width=0.075\textwidth]{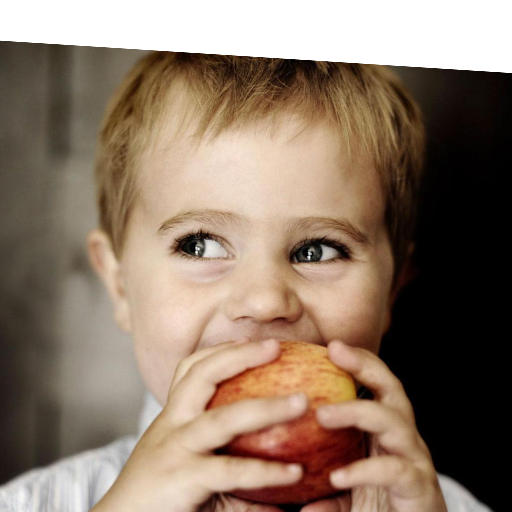}
\includegraphics[width=0.075\textwidth]{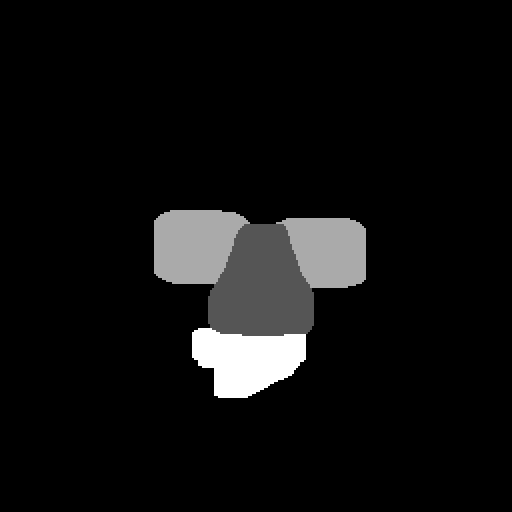}
\includegraphics[width=0.075\textwidth]{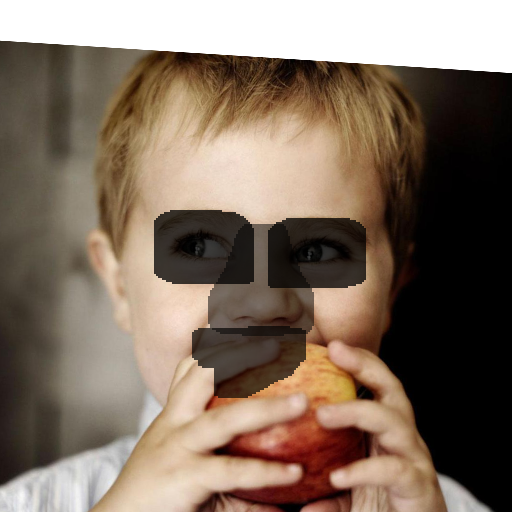}
\includegraphics[width=0.075\textwidth]{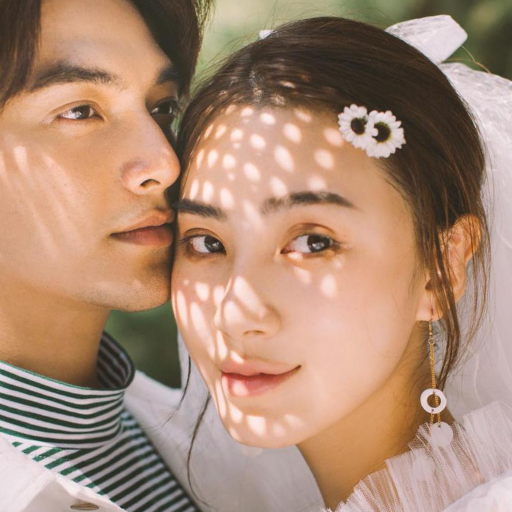}
\includegraphics[width=0.075\textwidth]{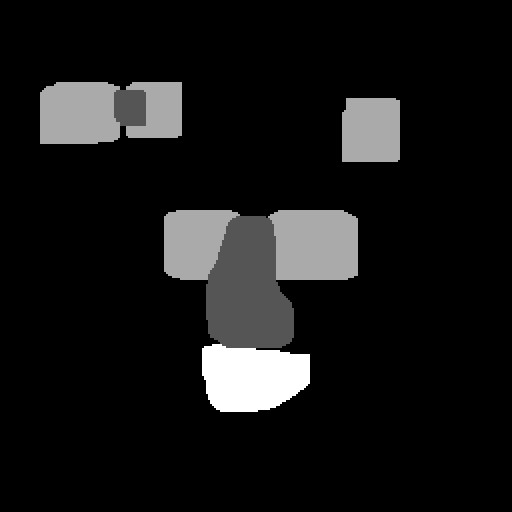}
\includegraphics[width=0.075\textwidth]{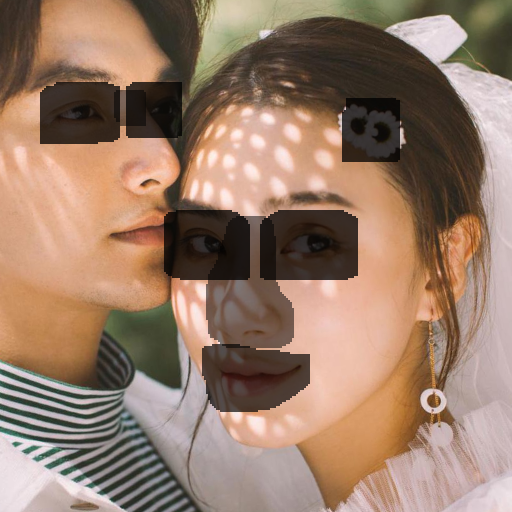}
\end{center}
\vspace{-0.15in}
\caption{Examples of face parsing masks.}
\label{fig:parsing_mask}
\end{figure}

We further \lyj{perform a comparison by replacing local discriminators with a single discriminator which uses a new channel containing face region information.}
Our experiment shows that the results of this ablation are worse than those by our method, e.g., with partial facial features missing or messy (Figure~\ref{fig:compare_addchannel}).
Also note that the face parsing masks are computed by an off-the-shelf face parsing network, with the parsed eyes/nose/lips regions dilated to make them cover the facial features. Some examples of the face parsing masks are shown in Figure~\ref{fig:parsing_mask}.
\lyj{The results show that} our system does not require accurate parsing masks.

\rev{The quantitative evaluation of the above ablation studies and comparisons are reported in Table~\ref{table:fid_ab}. 
The FID values of these ablation studies are worse than ours:
\begin{itemize}
    \item without style feature and style loss, the generated results are not of a uniform style, so the distance to each style is much larger than ours;
    \item without truncation loss, the FID also increases (worse).
\end{itemize}  
\noindent These results show that the ablated components (style feature, truncation loss) are essential for our model.
The comparison of replacing local discriminators with a single discriminator using a new channel has much larger FID value than ours, indicating a single discriminator using a new channel is harder to train, and our design of introducing local discriminators for important facial regions is more effective. Avg $\Delta$ shows the average difference between our method and each ablated version. }

\begin{table}[t]
  \centering
  \caption{\rev{Fr\'{e}chet Inception Distance (FID) of more ablation studies and comparisons. The FID values are computed between the set of generated APDrawings of each style and the collected true drawings of the corresponding style.}}
  \setlength\tabcolsep{4pt}
  {
  \begin{tabular}{c|ccc|c}
  \hline
  Methods & Style1 $\downarrow$ & Style2 $\downarrow$ & Style3 $\downarrow$ & Avg $\Delta$\\
  \hline
  w/o style feature & 114.3 & 122.1 & 111.5 & 20.90 \\
  w/o truncation loss & 81.7 & 120.1 & 99.2 & 5.27 \\
  replace $D_{l*}$ with a single $D$ & 93.6 & 152.8 & 124.0 & 28.40 \\
  Ours & \textbf{81.2} & \textbf{114.3} & \textbf{89.7} & /\\
  \hline
  \end{tabular}}
  \label{table:fid_ab}
\end{table}

\section{More Results}

\subsection{Material in the User Study}
\label{ssec:compare60}
In Section \exten{7.2} of the main paper, we compare our method with state-of-the-art methods in neural style transfer and image translation.
In Section \exten{7.3} of the main paper, we conduct a user study in which users sort the results of four methods (\exten{CycleGAN~\cite{ZhuPIE17}, ComboGAN~\cite{AnooshehATG18}, our conference version~\cite{YiLLR20}} and our method).
\rev{Each time, users compare different methods' results of a single style.
We denote 1 input photo and 4 generated drawings of a single style as a group.
In total, 60 groups are evaluated in this user study. 
Among them, 20 groups are for style1 comparison, 20 groups are for style2 and 20 groups are for style3. 
We show all 60 groups in Figures~\ref{fig:more_compare1}-\ref{fig:more_compare3-2}.
For a more comprehensive comparison, we show results of all the 3 styles for the multi-modal methods (ComboGAN, our conference and ours) and highlight the compared group in the user study in green boxes.}
Note that all these 60 groups are randomly chosen from the test set.
Our method outperforms the other three methods in most groups in terms of style similarity, face structure preservation and image visual quality.
The results of the user study summarized in Section~\exten{7.3} of the main paper also demonstrate the advantage of our method, where \exten{43.0\%} votes chose our method to be the best among the four methods\exten{, higher than the best vote percentages of the other \lyj{three} methods}.

\begin{figure}[t]
\centering
\footnotesize
\makebox[0.096\textwidth]{Input}
\makebox[0.096\textwidth]{\rev{APDrawingGAN++}}
\makebox[0.288\textwidth]{Ours(style1, 2, 3)}
\\
\normalsize
\includegraphics[width = 0.48\textwidth]{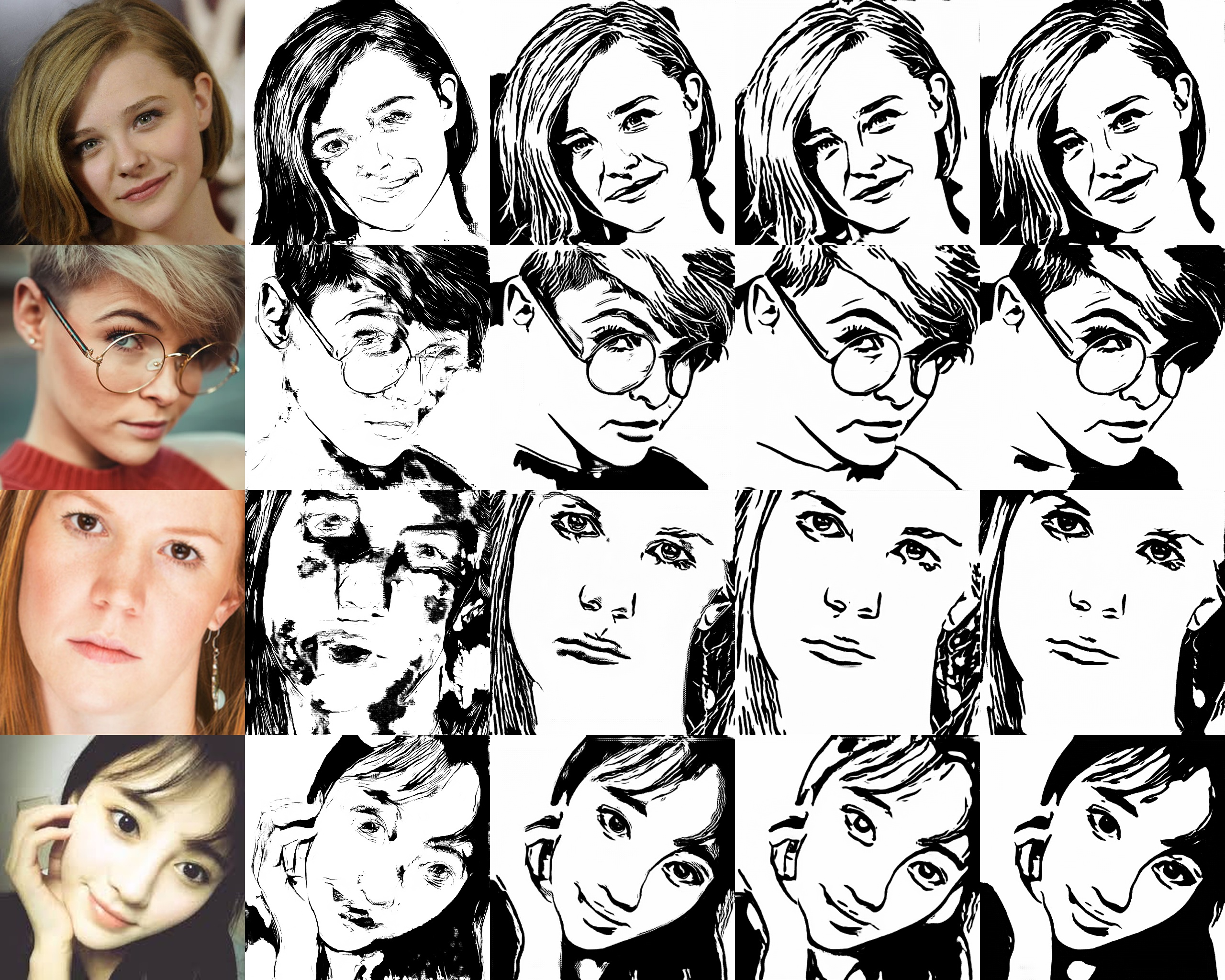}
\caption{Comparisons of APDrawingGAN++ and our method on challenging photos with arbitrary head orientation. From left to right: input photos, APDrawingGAN++ results, and our results (styles 1, 2, 3).}
\label{fig:compare_headtilt}
\end{figure}

\subsection{Comparison with \lyj{APDrawingGAN++}}
\label{ssec:apdrawinggan}

\lyj{APDrawingGAN++~\cite{YiLLR20pami}} is a deep neural network model specially designed for APDrawing generation by using a hierarchical structure and a distance transform loss.
However, this method requires \emph{paired} training data and cannot adapt well to face photos with unconstrained lighting in the wild due to the limited availability of paired training data. In comparison, our method only uses \emph{unpaired} training data, which makes it possible to include more challenging photos into the training set. Therefore, our method can generate high quality APDrawings for challenging photos under various conditions.
We compare the visual quality of \lyj{APDrawingGAN++} and our method using some challenging examples as illustrated in Figure \ref{fig:comp_apd++}.
These challenging examples include unconventional lighting conditions (1st-\exten{4}th rows), unconventional expression or taking accessories like sunglasses (\exten{5th-7th} rows), or blurry looking (\exten{8th-9th} rows, zoom in to check).
\lyj{APDrawingGAN++} generates messy results for these challenging photos, while our method generates high-quality APDrawings with much better visual effect.

Moreover, \lyj{APDrawingGAN++} uses a hierarchical network structure that feeds local rectangle regions around eyes, nose and mouth centers into local generators and discriminators. This setting cannot tolerate a large head tilt and requires that its input photos are in the upright orientation (i.e., the photo needs to be rotated so that the two eyes are on a horizontal line). Then the local regions of eyes, nose and mouth can be covered by rectangle regions.
In comparison, although our model also has local discriminators, we use face masks (obtained from a face parsing network~\cite{GuBY0WY19}), and the inputs to local discriminators are the masked eyes, nose, mouth regions.
Therefore our method does not need the input images to be adjusted into the upright orientation.
Comparisons of \lyj{APDrawingGAN++ and our method} on face photos with arbitrary head orientation are shown in Figure~\ref{fig:compare_headtilt}.
The results show that \lyj{APDrawingGAN++} often generates messy results and some boundaries of rectangle local regions are clearly visible, whereas our results are clean and have good visual quality.

\subsection{More Tests on the CelebAMask-HQ Dataset}
\label{ssec:more_test}
In the main paper, we test our model on photos collected from Internet.
Here, we further test our method on photos from the CelebAMask-HQ Dataset~\cite{abs-1907-11922}.
The results are summarized in Figure~\ref{fig:more_test},
\lyj{showing that} our method generates high quality results with good image and line quality on the CelebAMask-HQ Dataset.

\begin{figure*}[t]
\centering
\includegraphics[width = 1.0\textwidth]{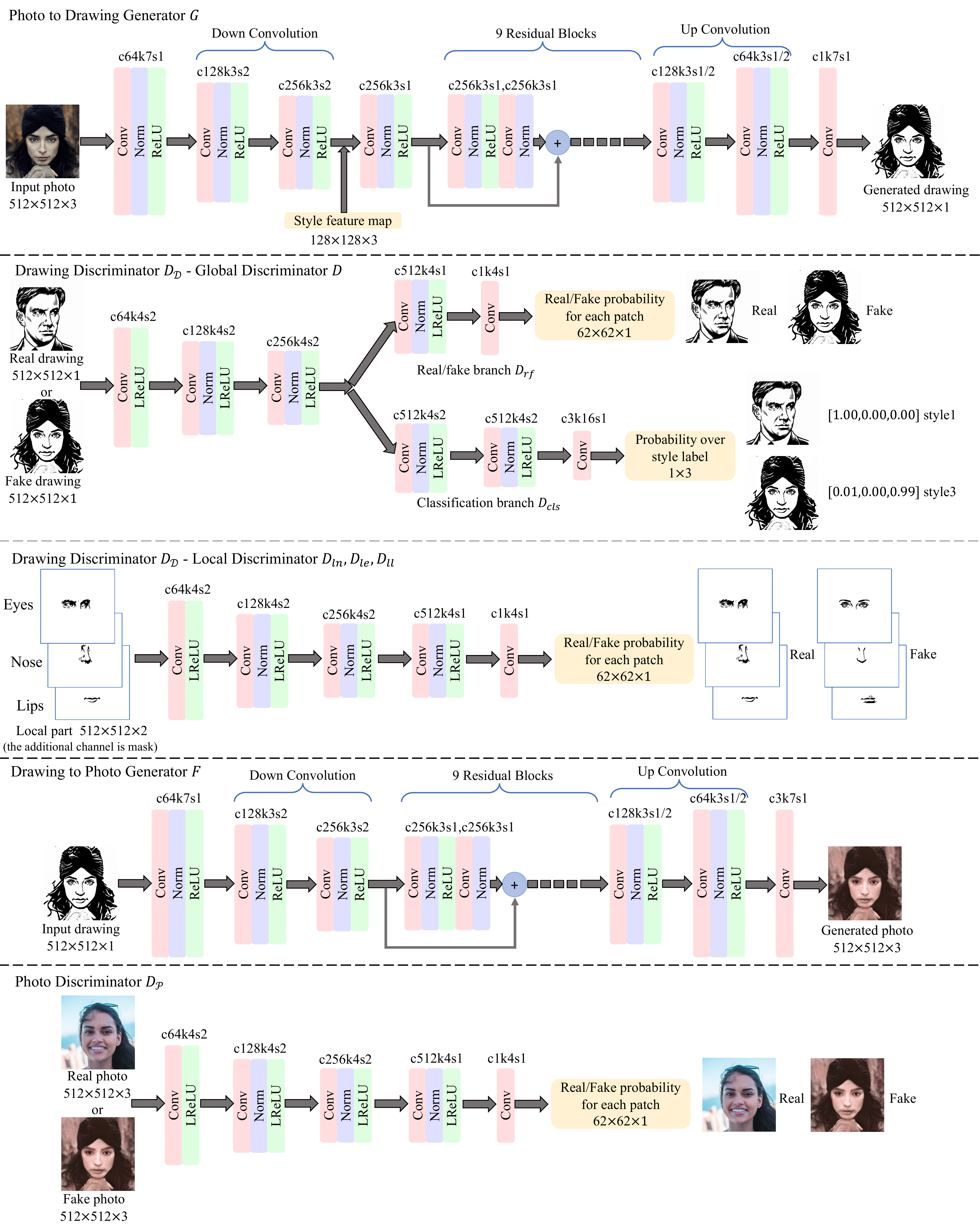}
\caption{\exten{Detailed network architecture of our model. We denote the output channel number as $c$, convolution kernel size as $k$, and stride in a convolution layer as $s$.
`Norm' means the instance normalization layer, and `LReLU' means the leaky ReLU with $\alpha=0.2$.}}
\label{fig:network_details}
\end{figure*}

\begin{figure*}[t]
\centering
\small
\makebox[0.64in]{Input}
\makebox[0.64in]{CycleGAN}
\makebox[1.91in]{ComboGAN (style 1, 2, 3)}
\makebox[1.91in]{Ours-pre (style 1, 2, 3)}
\makebox[1.91in]{Ours (style 1, 2, 3)}\\
\normalsize
\includegraphics[width = 1\textwidth]{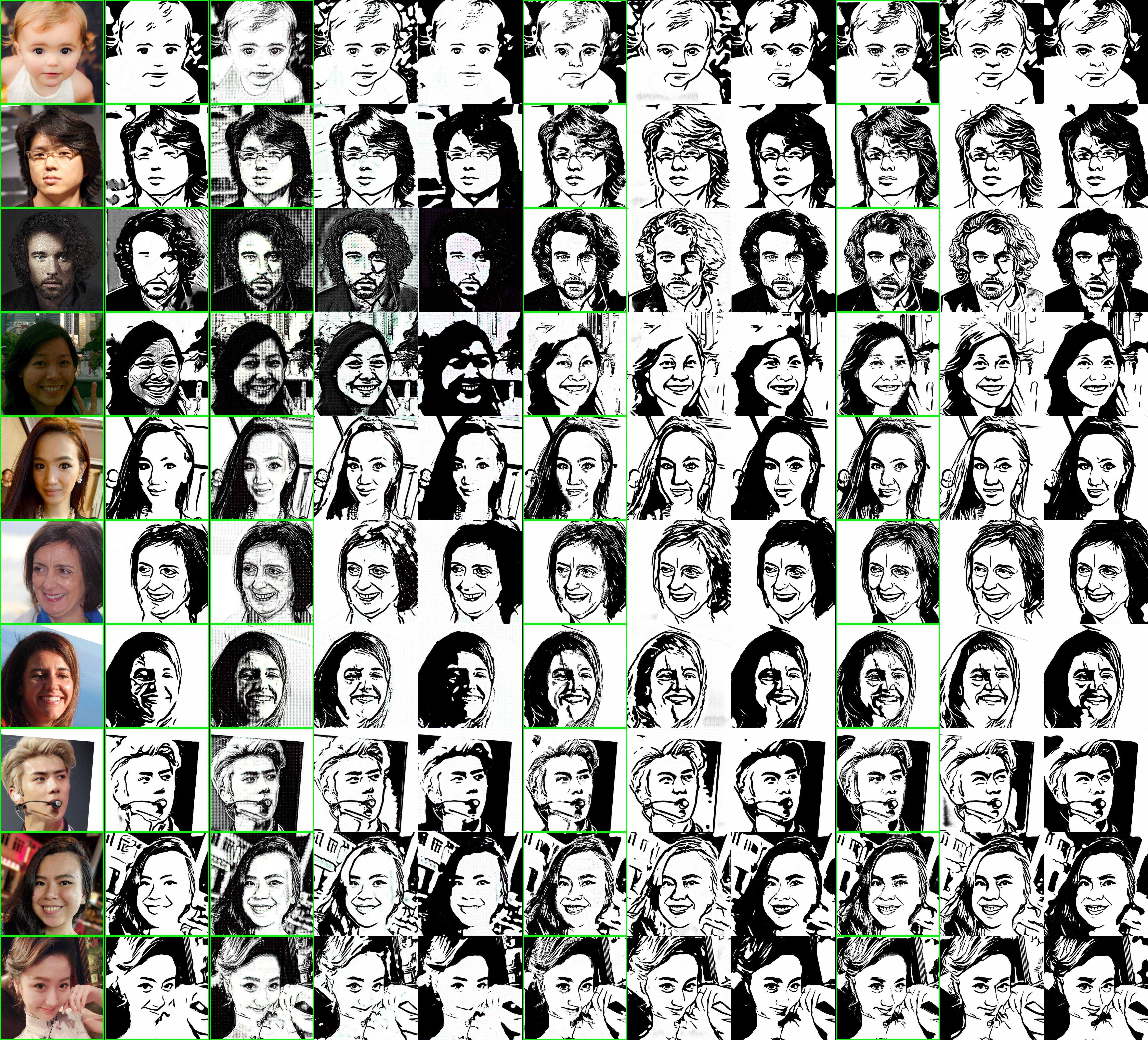}
\caption{\rev{More qualitative comparisons (user study material). From left to right: input face photos, CycleGAN~\cite{ZhuPIE17} results, ComboGAN~\cite{AnooshehATG18} results (style 1, 2, 3), results of our conference version (Ours-pre)~\cite{YiLLR20} (style 1, 2, 3), and our results (style 1, 2, 3). In the user study, users compared each time the results of a single style. 60 groups are evaluated and there are 20 groups for each style. We show results of all the 3 styles and highlight the compared group in green boxes.}}
\label{fig:more_compare1}
\end{figure*}

\begin{figure*}[t]
\centering
\small
\makebox[0.64in]{Input}
\makebox[0.64in]{CycleGAN}
\makebox[1.91in]{ComboGAN (style 1, 2, 3)}
\makebox[1.91in]{Ours-pre (style 1, 2, 3)}
\makebox[1.91in]{Ours (style 1, 2, 3)}\\
\normalsize
\includegraphics[width = 1\textwidth]{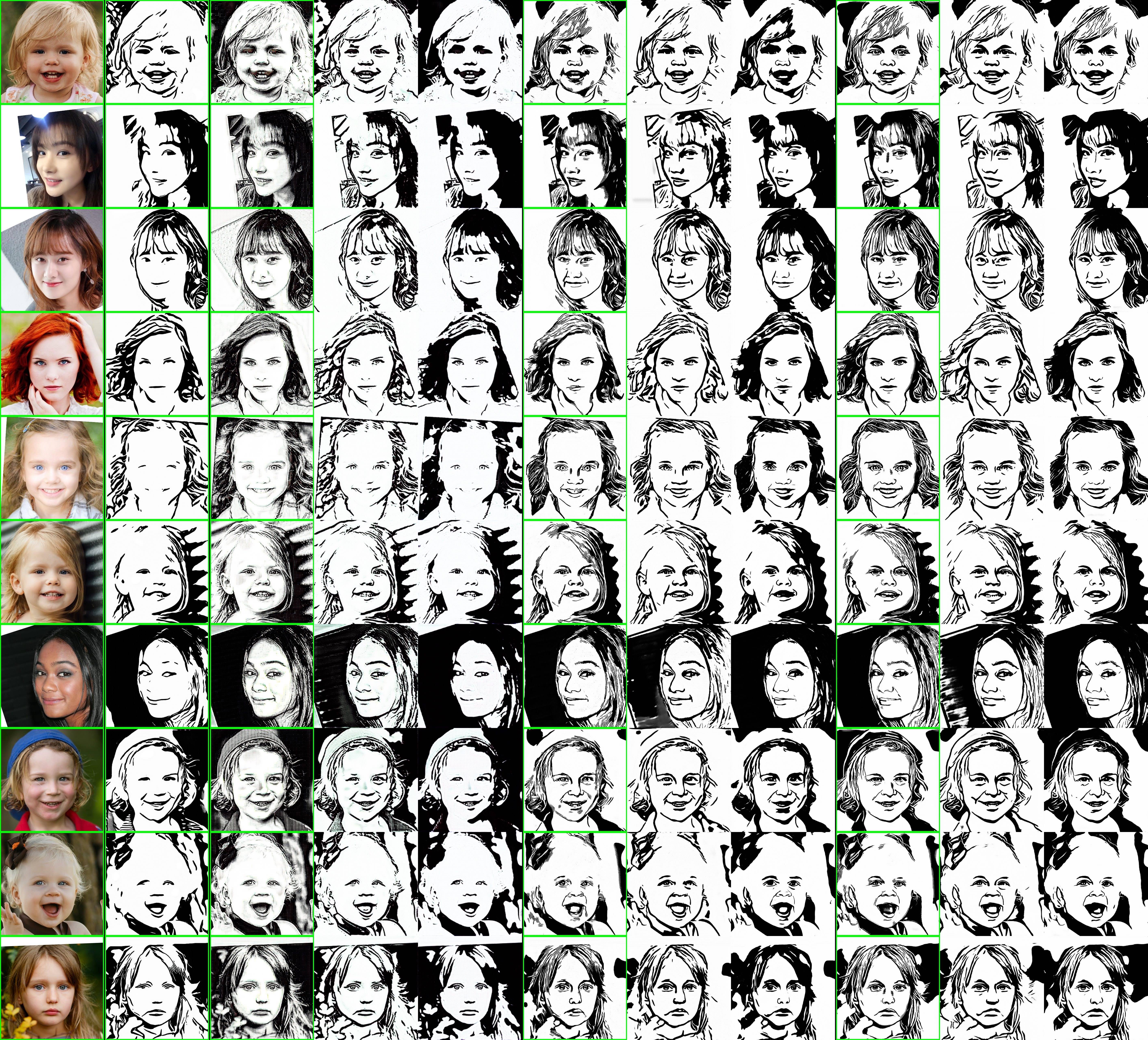}
\caption{\rev{More qualitative comparisons (user study material). From left to right: input face photos, CycleGAN~\cite{ZhuPIE17} results, ComboGAN~\cite{AnooshehATG18} results (style 1, 2, 3), results of our conference version (Ours-pre)~\cite{YiLLR20} (style 1, 2, 3), and our results (style 1, 2, 3). In the user study, each time users compared results of a single style. 60 groups are evaluated and there are 20 groups for each style. We show results of all the 3 styles and highlight the compared group in green boxes.}}
\label{fig:more_compare1-2}
\end{figure*}

\begin{figure*}[t]
\centering
\small
\makebox[0.64in]{Input}
\makebox[0.64in]{CycleGAN}
\makebox[1.91in]{ComboGAN (style 1, 2, 3)}
\makebox[1.91in]{Ours-pre (style 1, 2, 3)}
\makebox[1.91in]{Ours (style 1, 2, 3)}\\
\normalsize
\includegraphics[width = 1\textwidth]{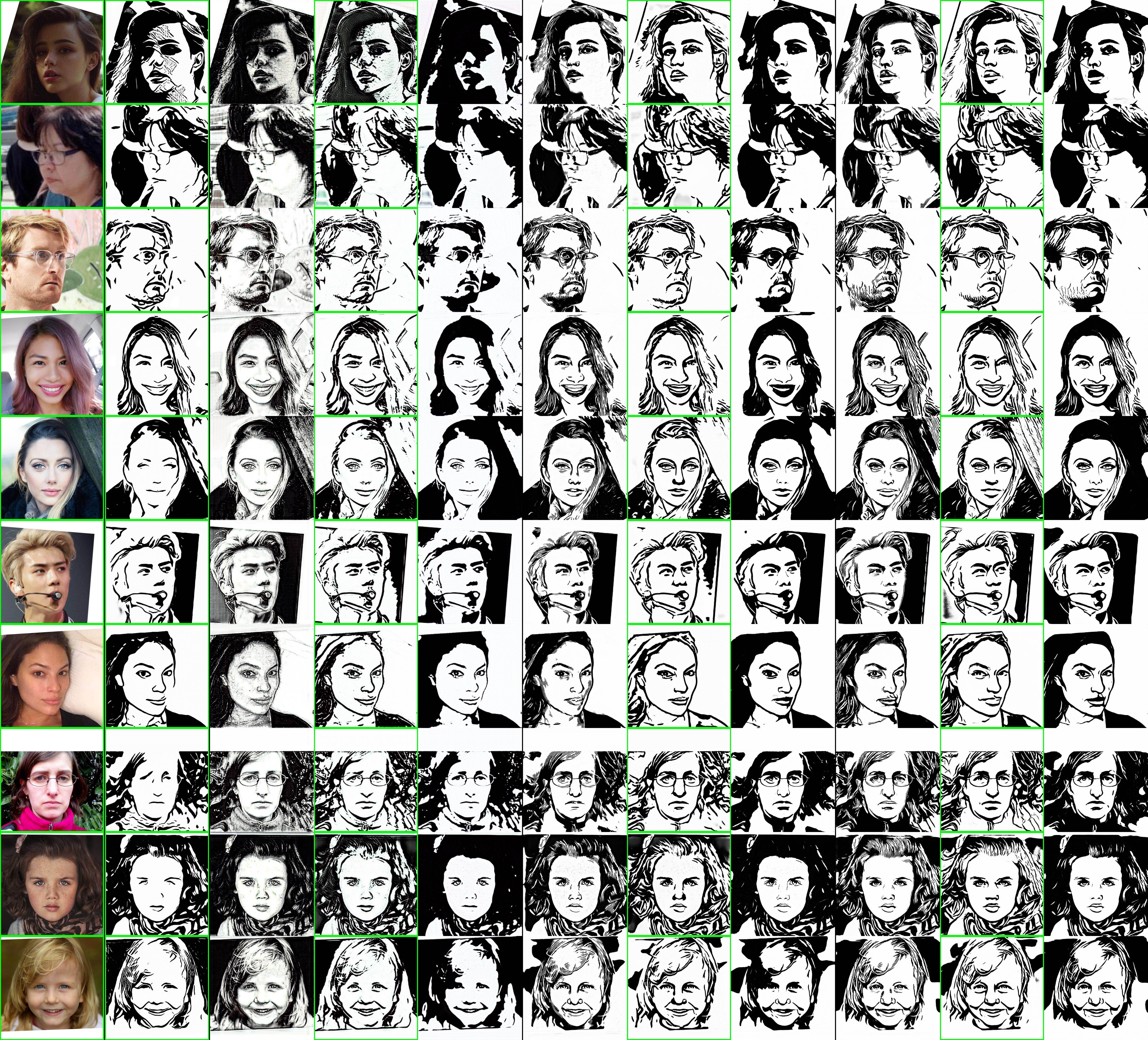}
\caption{\rev{More qualitative comparisons (user study material). From left to right: input face photos, CycleGAN~\cite{ZhuPIE17} results, ComboGAN~\cite{AnooshehATG18} results (style 1, 2, 3), results of our conference version (Ours-pre)~\cite{YiLLR20} (style 1, 2, 3), and our results (style 1, 2, 3). In the user study, users compared each time the results of a single style. 60 groups are evaluated and there are 20 groups for each style. We show results of all the 3 styles and highlight the compared group in green boxes.}}
\label{fig:more_compare2}
\end{figure*}

\begin{figure*}[t]
\centering
\small
\makebox[0.64in]{Input}
\makebox[0.64in]{CycleGAN}
\makebox[1.91in]{ComboGAN (style 1, 2, 3)}
\makebox[1.91in]{Ours-pre (style 1, 2, 3)}
\makebox[1.91in]{Ours (style 1, 2, 3)}\\
\normalsize
\includegraphics[width = 1\textwidth]{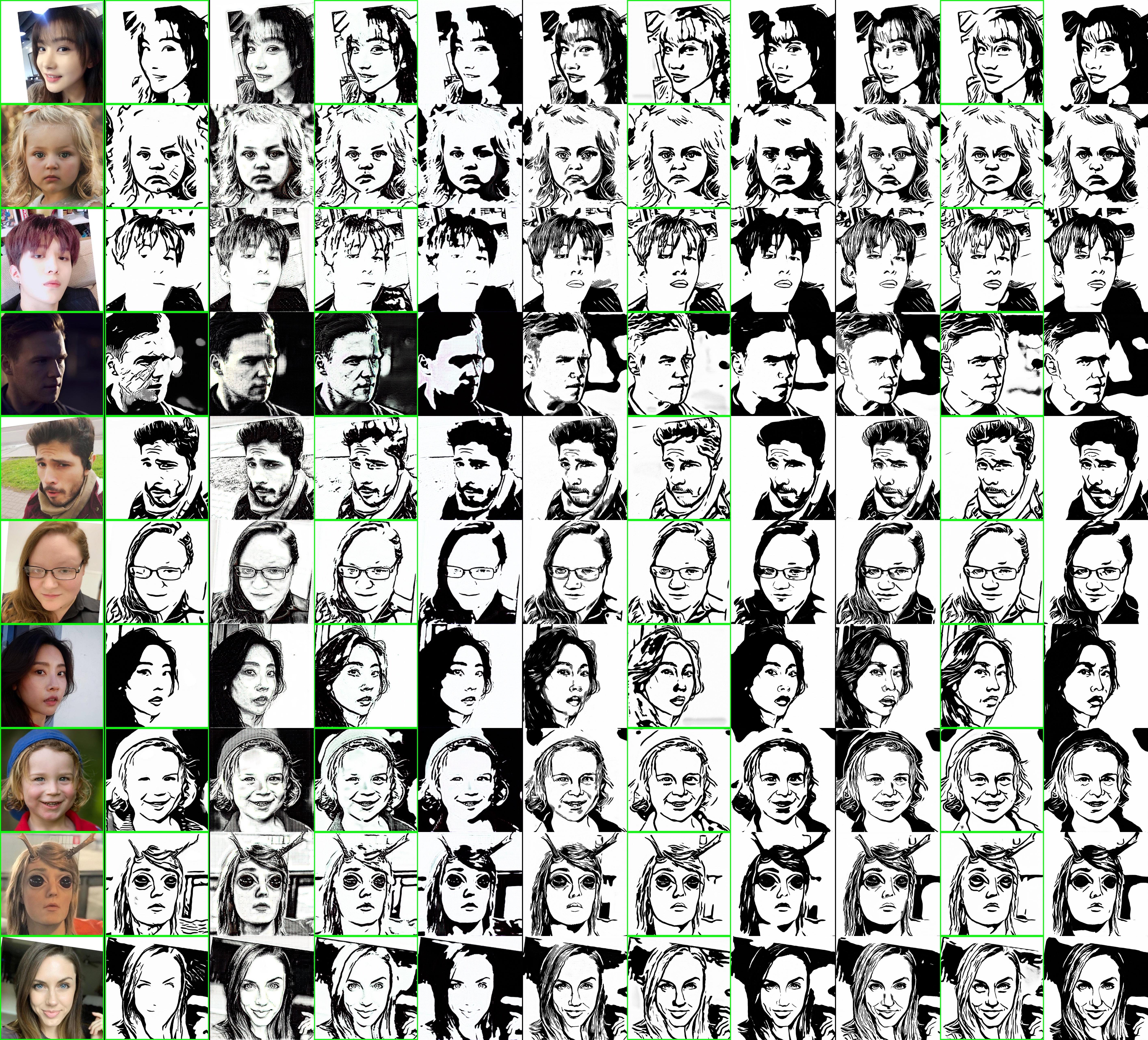}
\caption{\rev{More qualitative comparisons (user study material). From left to right: input face photos, CycleGAN~\cite{ZhuPIE17} results, ComboGAN~\cite{AnooshehATG18} results (style 1, 2, 3), results of our conference version (Ours-pre)~\cite{YiLLR20} (style 1, 2, 3), and our results (style 1, 2, 3). In the user study, users compared each time the results of a single style. 60 groups are evaluated and there are 20 groups for each style. We show results of all the 3 styles and highlight the compared group in green boxes.}}
\label{fig:more_compare2-2}
\end{figure*}

\begin{figure*}[t]
\centering
\small
\makebox[0.64in]{Input}
\makebox[0.64in]{CycleGAN}
\makebox[1.91in]{ComboGAN (style 1, 2, 3)}
\makebox[1.91in]{Ours-pre (style 1, 2, 3)}
\makebox[1.91in]{Ours (style 1, 2, 3)}\\
\normalsize
\includegraphics[width = 1\textwidth]{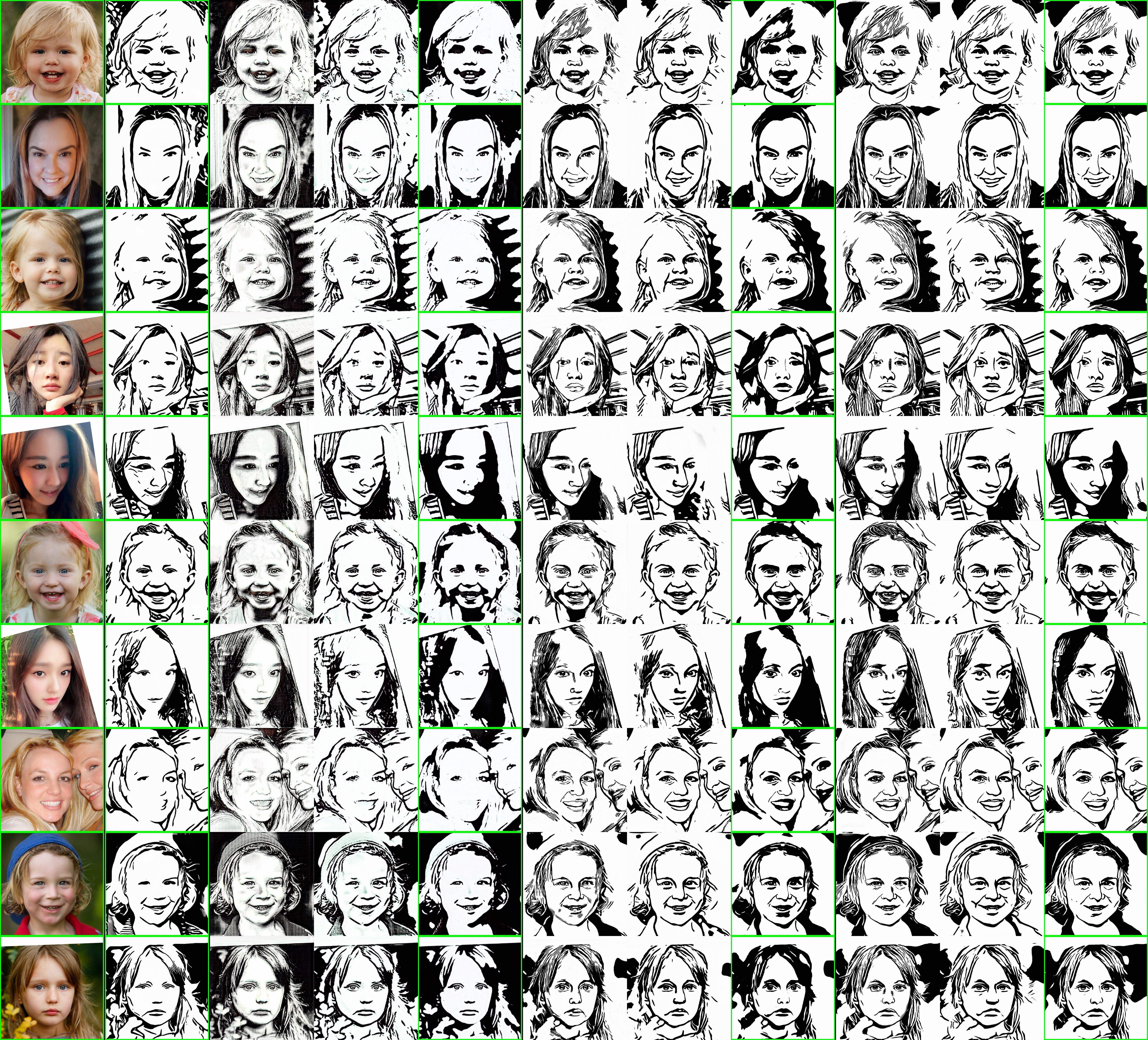}
\caption{\rev{More qualitative comparisons (user study material). From left to right: input face photos, CycleGAN~\cite{ZhuPIE17} results, ComboGAN~\cite{AnooshehATG18} results (style 1, 2, 3), results of our conference version (Ours-pre)~\cite{YiLLR20} (style 1, 2, 3), and our results (style 1, 2, 3). In the user study, each time users compared results of a single style. 60 groups are evaluated and there are 20 groups for each style. We show results of all the 3 styles and highlight the compared group in green boxes.}}
\label{fig:more_compare3}
\end{figure*}

\begin{figure*}[t]
\centering
\small
\makebox[0.64in]{Input}
\makebox[0.64in]{CycleGAN}
\makebox[1.91in]{ComboGAN (style 1, 2, 3)}
\makebox[1.91in]{Ours-pre (style 1, 2, 3)}
\makebox[1.91in]{Ours (style 1, 2, 3)}\\
\normalsize
\includegraphics[width = 1\textwidth]{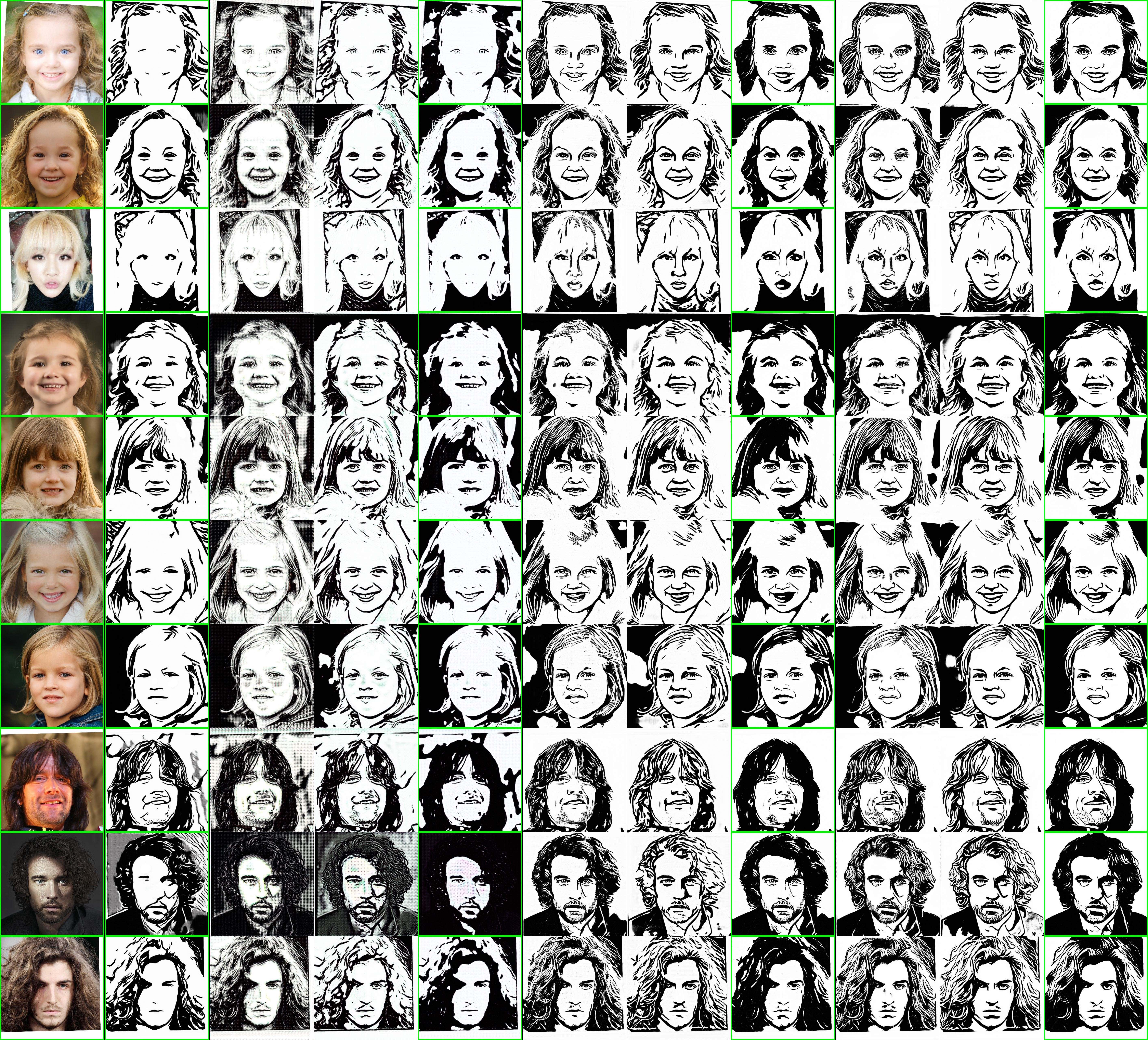}
\caption{\rev{More qualitative comparisons (user study material). From left to right: input face photos, CycleGAN~\cite{ZhuPIE17} results, ComboGAN~\cite{AnooshehATG18} results (style 1, 2, 3), results of our conference version (Ours-pre)~\cite{YiLLR20} (style 1, 2, 3), and our results (style 1, 2, 3). In the user study, each time users compared results of a single style. 60 groups are evaluated and there are 20 groups for each style. We show results of all the 3 styles and highlight the compared group in green boxes.}}
\label{fig:more_compare3-2}
\end{figure*}

\begin{figure*}[t]
\centering
\footnotesize
\makebox[0.81in]{Input}
\makebox[0.81in]{\rev{APDrawingGAN++}}
\makebox[0.81in]{Ours(style1)}
\makebox[0.81in]{Ours(style2)}
\makebox[0.81in]{Ours(style3)}
\\
\normalsize
\includegraphics[width = 0.62\textwidth]{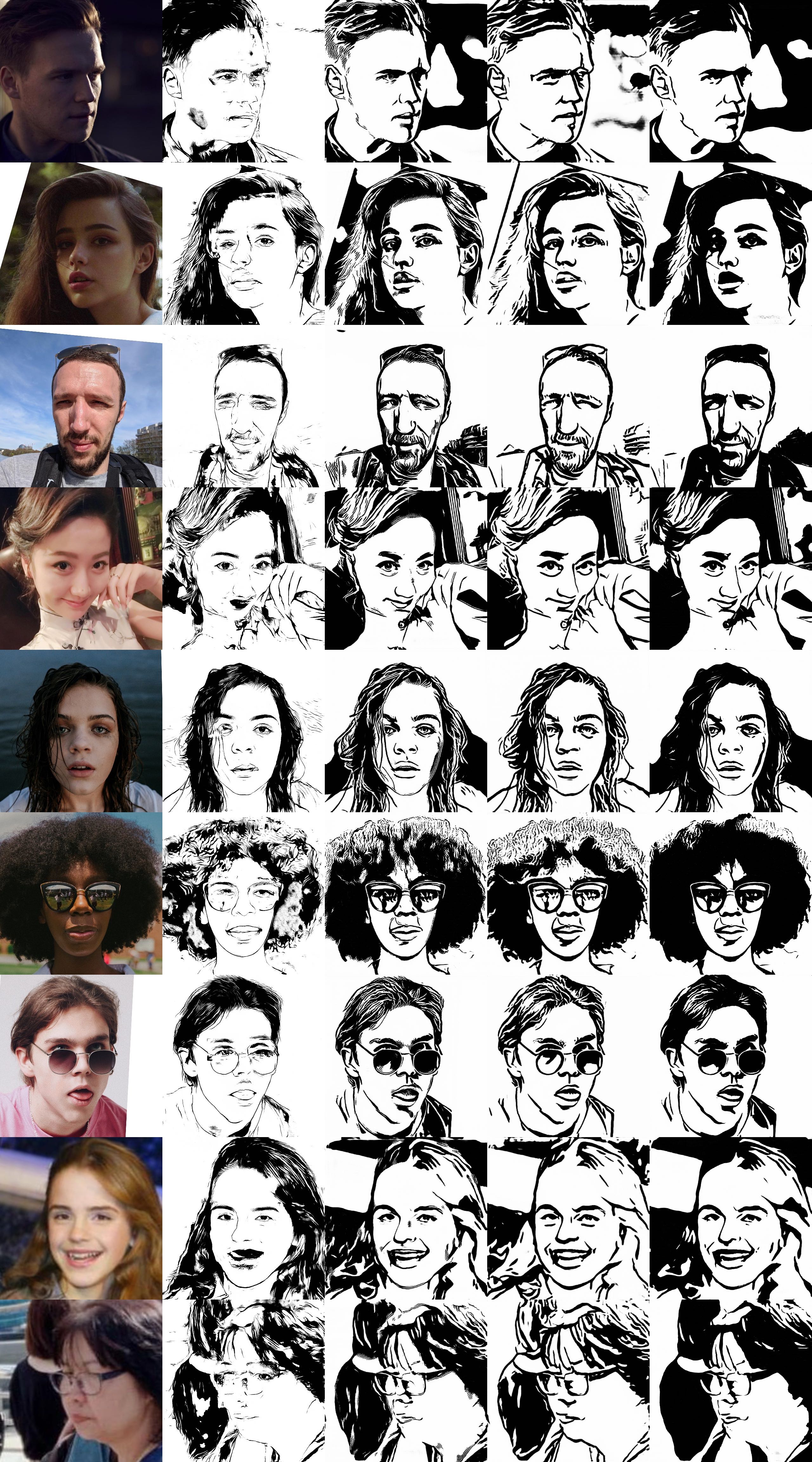}
\caption{Comparison of APDrawingGAN++~\cite{YiLLR20pami} and our method on face photos under some challenging situations. From left to right: input face photos, APDrawingGAN++~\cite{YiLLR20pami} results, our results (style1), our results (style2), our results (style3). 
The face photos in the 5-7th rows are from NPRportrait1.0 Benchmark~\cite{rosin2021nprportrait}.
\rev{The face photo in the 8th row is from LFW Dataset~\cite{LFWTech}.}}
\label{fig:comp_apd++}
\end{figure*}

\begin{figure*}[t]
\centering
\small
\makebox[0.81in]{Input}
\makebox[0.81in]{Ours(style1)}
\makebox[0.81in]{Ours(style2)}
\makebox[0.81in]{Ours(style3)}
\makebox[0.81in]{Input}
\makebox[0.81in]{Ours(style1)}
\makebox[0.81in]{Ours(style2)}
\makebox[0.81in]{Ours(style3)}
\\
\normalsize
\includegraphics[width = 0.98\textwidth]{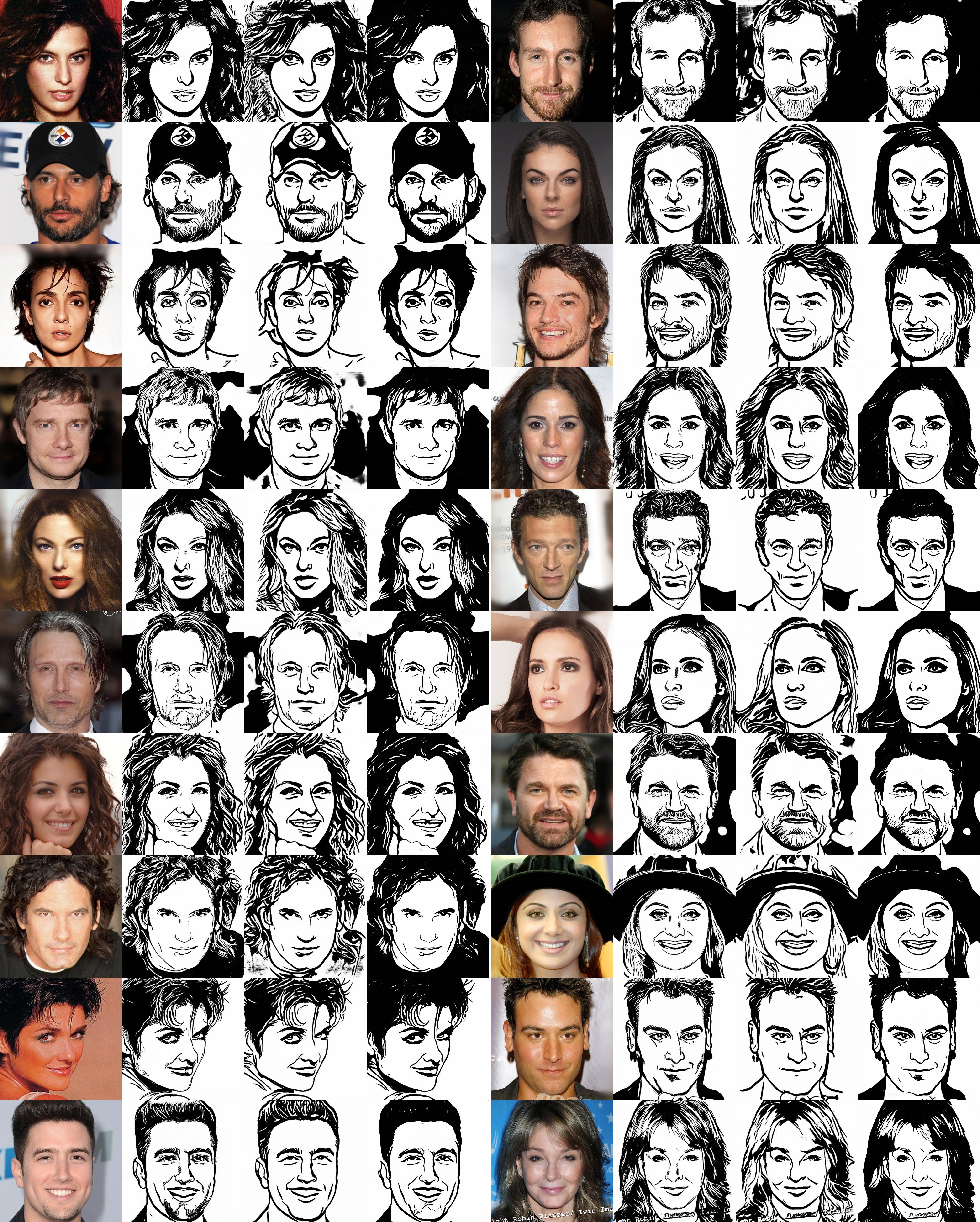}
\caption{More test results on CelebAMask-HQ Dataset~\cite{abs-1907-11922}. From left to right: input face photos, our results (style1), our results (style2), our results (style3), input face photos, our results (style1), our results (style2), our results (style3).}
\label{fig:more_test}
\end{figure*}



\ifCLASSOPTIONcaptionsoff
  \newpage
\fi



\bibliographystyle{IEEEtran}
\bibliography{IEEEabrv,jrnl}
%
%
%

%

\begin{IEEEbiography}[{\includegraphics[width=1in,height=1.25in,clip,keepaspectratio]{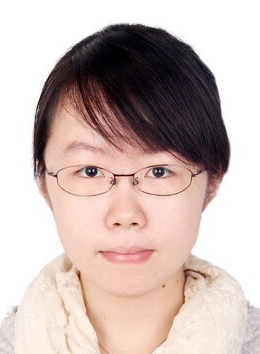}}]{Ran Yi}
is currently an Assistant Professor with Department of Computer Science and Engineering, Shanghai Jiao Tong University, China.
She received her B.Eng. and Ph.D degree from Tsinghua University, China, in 2016 and 2021 respectively.
Her research interests include computer vision, computer graphics and machine intelligence.
\end{IEEEbiography}

\begin{IEEEbiography}[{\includegraphics[width=1in,height=1.25in,clip,keepaspectratio]{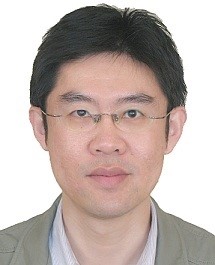}}]{Yong-Jin Liu}
is a Professor with the Department of Computer Science and Technology, Tsinghua University, China.
He received the BEng degree from Tianjin University, China, in 1998, and the PhD degree from the Hong Kong University of Science and Technology, Hong Kong, China, in 2004.
His research interests include computational geometry, computer graphics and computer vision. He is a senior member of the IEEE. For more information, visit \url{https://cg.cs.tsinghua.edu.cn/people/~Yongjin/Yongjin.htm}
\end{IEEEbiography}

\begin{IEEEbiography}[{\includegraphics[width=1in,height=1.25in,clip,keepaspectratio]{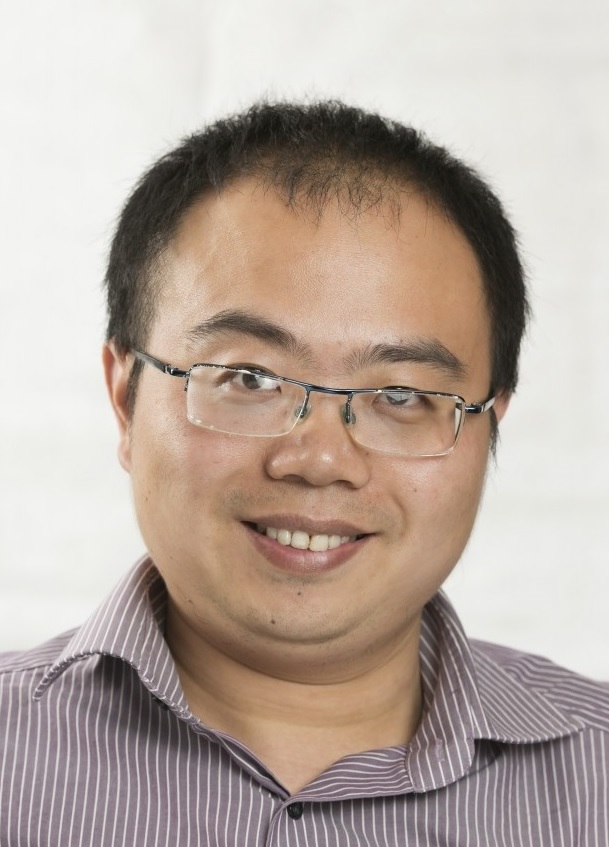}}]{Yu-Kun Lai}
is a Professor at School of Computer Science and Informatics,
Cardiff University, UK. He received his B.S and PhD degrees in Computer Science from Tsinghua University, in 2003 and 2008 respectively.
His research interests include computer graphics, computer vision, geometric modeling and image processing.
For more information, visit \url{https://users.cs.cf.ac.uk/Yukun.Lai/}
\end{IEEEbiography}

\begin{IEEEbiography}[{\includegraphics[width=1in,height=1.25in,clip,keepaspectratio]{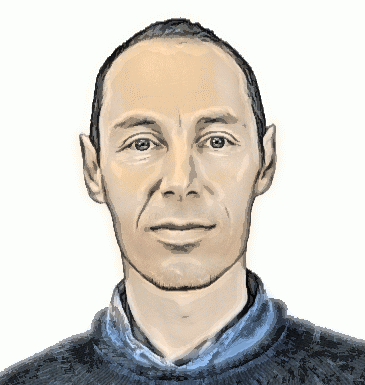}}]{Paul L. Rosin}
is a Professor at School of Computer Science and Informatics,
Cardiff University, UK.
Previous posts include lecturer at the Department of Information Systems and Computing, Brunel University London, UK, research scientist at the Institute for Remote Sensing Applications, Joint Research Centre, Ispra, Italy, and lecturer at Curtin University of Technology, Perth, Australia.
His research interests include low level image processing, performance evaluation, shape analysis, facial analysis, cellular automata, non-photorealistic rendering and cultural heritage. For more information, visit \url{http://users.cs.cf.ac.uk/Paul.Rosin/}
\end{IEEEbiography}







\end{document}